\documentclass[11pt]{article}

\usepackage[preprint]{acl}

\usepackage{times}
\usepackage{latexsym}

\usepackage[T1]{fontenc}
\usepackage[utf8]{inputenc}

\usepackage{microtype}

\usepackage{inconsolata}

\usepackage{graphicx}
\graphicspath{{imgs/}} % Path for images' folder
\usepackage{eso-pic} % For the background picture on the title page
\usepackage{subfig} % Numbered and caption subfigures using \subfloat
\usepackage{caption} % Coloured captions
\usepackage{transparent}

\usepackage{amsmath}
\usepackage{amsthm}
\usepackage{bm}
\usepackage[overload]{empheq}  % For braced-style systems of equations

\usepackage{tabularx}
\usepackage{longtable} % tables that can span several pages
\usepackage{colortbl}

\usepackage{booktabs}
\usepackage{multirow}
\usepackage{amssymb}
\usepackage{enumitem}

\title{Just how sure are you? Improving Verbalized Uncertainty Calibration in Medical VQA}

\author{
 \textbf{Eren Senoglu\textsuperscript{1}},
 \textbf{Federico Toschi\textsuperscript{1}},
 \textbf{Nicolò Brunello\textsuperscript{1}},
 \textbf{Andrea Sassella\textsuperscript{1}},\\
 \textbf{Mark J. Carman\textsuperscript{1}}
\\
\\
  \textsuperscript{1}Politecnico di Milano
}

\begin{document}
\maketitle

\begin{abstract}
Multimodal large language models (MLLMs) applied to Medical Visual Question Answering (VQA) tend to produce overconfident outputs regardless of actual correctness, and existing verbalized confidence calibration methods, developed primarily for text-only LLMs, do not account for the multimodal nature of medical image understanding. 

This work proposes a training-based framework that fine-tunes MLLMs to improve their calibration using a composite 
loss function combining a Brier-style calibration term, an anchor regularizer that prevents confidence collapse toward extreme values, a contrastive image-text alignment term, and a KL-based model stabilization term. The alignment signal is derived from a $2\times2$ factorial perturbation design that crosses image presence with text integrity, probing the model's reliance on visual modality input versus language priors. Finally, a top-$k$ KL divergence regularizer is used to protect the model’s answering ability during fine-tuning. 

Across three Medical VQA benchmarks and two architectures (MedGemma-4B-IT and Qwen2-VL-7B-Instruct), our method reduces calibration error by 60\% or more, and improves discrimination by 26\% or more, while preserving predictive accuracy. On average across benchmarks, the technique outperforms prompting-based, sampling-based, and training-based approaches, and ablation experiments confirm that each component of the loss function is indeed necessary for improving the calibration. All code for the experiments is publicly available.\footnote{\url{https://github.com/ErenSenoglu/Verbalized-Uncertainty-Calibration-for-MedVQA}}

%Multimodal large language models (MLLMs) applied to Medical Visual Question Answering (VQA) tend to produce overconfident outputs regardless of actual correctness, and existing verbalized confidence calibration methods, developed primarily for text-only LLMs, do not account for the multimodal nature of medical image understanding. 

%This work proposes a training-based calibration framework built on a $2\times2$ factorial perturbation design that crosses image availability with text integrity, a composite loss that combines Brier-based calibration with an anchor regularizer and a contrastive alignment term derived from the perturbation conditions, and a top-$k$ KL divergence regularizer that preserves answering ability during confidence training. 

%Across three Medical VQA benchmarks and two architectures (MedGemma-4B-IT and Qwen2-VL-7B-Instruct), our method reduces calibration error by more than 50\% relative to the base model, improves discrimination by 27\%, and outperforms all compared baselines on average across both combined calibration and discrimination metrics.
%All code for the experiments is publicly available.\footnote{\url{https://anonymous.4open.science/r/Verbalized-Calibration-for-MedVQA-F6E2/}}
\end{abstract}

\section{Introduction}
\label{sec:introduction}

Medical AI systems have progressed from black-box models that produce predictions without explanation, to MLLMs that generate interpretable reasoning traces alongside their answers. A natural next step in this progression is calibrated self-assessment: a model that not only explains its reasoning but also communicates how much it trusts its own answer. Recent benchmarks show that MLLMs achieve competitive performance on medical knowledge tasks~\cite{sellergren2025medgemma} and, in some settings, match or exceed physician-level diagnostic accuracy~\cite{sheng2026nejm, nori2025sequential}, driving interest in clinical applications such as report drafting, case summarization, and decision support. However, these models consistently express high confidence even when their answers are incorrect, limiting their reliability in clinical use.

A clinician who receives a model-generated suggestion must manually verify the reasoning and check whether it is grounded in the image, a process that negates the efficiency gains the system is meant to provide. A more practical paradigm is one where the model communicates its own uncertainty, analogous to how a human colleague qualifies a recommendation, so that the user can focus verification effort on the cases the model itself flags as uncertain.

Existing verbalized confidence calibration methods have been developed for text-only LLMs and do not account for the multimodal nature of medical image understanding. In this work, we propose a training-based framework for verbalized confidence calibration in Medical VQA, built on three contributions: (1) a $2\times2$ factorial perturbation design that probes the model's reliance on visual evidence versus language priors, (2) a composite calibration loss that combines Brier-based calibration with a contrastive alignment term derived from the perturbation conditions, and (3) a top-$k$ KL divergence regularizer that preserves the answer distribution during confidence training. Across three benchmarks spanning different accuracy regimes and two model architectures, our method outperforms all compared baselines on average ECE, Brier Score, and AUROC.

\section{Related Work}
\label{sec:related_work}

Confidence estimation for LLMs falls into three broad families. Token-probability methods treat output logits directly as confidence scores, using sequence-level perplexity~\cite{malinin2021uncertainty} or semantic weighting of answer-relevant tokens~\cite{bakman2024mars}. Verbalized confidence methods prompt the model to express certainty in natural language~\cite{tian2023justask}. Consistency-based methods aggregate multiple sampled responses to estimate uncertainty through voting~\cite{wang2023selfconsistency}, semantic clustering~\cite{kuhn2023semantic}, or perturbation-based variance~\cite{zhang2024vluncertainty}. Top-K Sampling~\cite{xiong2024canllms} generates multiple answers per question and averages their confidence scores, while SteerConf~\cite{zhou2025steerconf} prompts the model at multiple steering levels and aggregates the responses into a composite score. Beyond these, internal-representation approaches extract confidence signals from hidden states or intermediate layers~\cite{xiao2025eagle, padhi2025harmony, liu2024actcab}, and architectural alternatives such as LARS~\cite{yaldiz2024lars} attach dedicated confidence heads to the model. Pre-trained models exhibit calibrated self-knowledge~\cite{kadavath2022language}, but token probabilities degrade after RLHF alignment~\cite{tian2023justask}, hidden-state methods require white-box access to model internals, and consistency methods require multiple inference passes at test time. Among these families, verbalized confidence is the most practical for deployment, as it requires no access to model internals and produces human-interpretable outputs, but the resulting confidence scores are poorly calibrated without further training~\cite{groot2024overconfidence, leng2024taming, xiao2025restoring}.

Calibrating verbalized confidence has therefore become an active research direction. \textbf{Supervision-based} (SFT) methods construct proxy targets through token-probability distillation~\cite{lin2022teaching}, post-fine-tuning self-assessment~\cite{chaudhry2024finetuning}, systematic question-context mutations~\cite{han2024lepe}, or clustering of sampled reasoning chains~\cite{xu2024sayself}. \textbf{Reward-based RL} methods instead optimize proper scoring rules directly: Rewarding Doubt~\cite{stangel2025rewardingdoubt} and LoVeC~\cite{zhang2025lovec} use logarithmic scoring, while ReCalibrate~\cite{yang2025recalibrate} uses the Brier score. ConfTuner~\cite{li2025conftuner} bridges both directions by replacing the SFT cross-entropy loss with a tokenized Brier score, computing the expected squared error under the model's probability distribution over confidence tokens. Unlike standard cross-entropy, which treats all token mismatches equally, this formulation naturally encodes ordinal structure: the penalty for predicting confidence 3 when the correct value is 2 is smaller than for predicting 7, achieving the theoretical guarantees of proper scoring rules without requiring policy optimization. To preserve answering ability during confidence training, it regularizes the answer tokens with a cross-entropy loss computed on the model's own outputs.

Despite their differences in training signal and algorithm, these methods share two limitations: they all operate on text-only LLMs without accounting for the visual modality, and they train on a single input condition, so the model never observes what happens when evidence sources are removed or disrupted. To the best of our knowledge, no existing method applies training-based calibration to multimodal models in the medical domain: multimodal work applies perturbation-based fine-tuning to object-level detection in general domains~\cite{zhao2025csp} or remains prompt-based~\cite{xuan2024vlmcalibration}, and medical methods rely on prompt engineering~\cite{kriz2025prompt4trust} rather than fine-tuning.

%-----------------------------------------------------------------------------
% METHODOLOGY
%-----------------------------------------------------------------------------
\section{Methodology}
\label{sec:methodology}

We formalize the task as follows. Given a medical image $\mathbf{x}$, a question $q$, and $A$ answer options $\mathcal{O}$, the model generates a reasoning trace $r$, an answer $a \in \mathcal{O}$, and a verbalized confidence score $c$ drawn from a discrete integer scale, normalized to $\hat{c} = c/N_A \in [0,1]$. We adopt an answer-first, confidence-second generation order, which prior work shows produces better calibrated outputs~\cite{li2025conftuner, zhou2025steerconf}. We evaluate confidence quality along two axes: \textbf{calibration} (stated confidence matches empirical correctness probability) and \textbf{discrimination} (correct predictions receive higher confidence than incorrect ones), using Expected Calibration Error (ECE)~\cite{guo2017calibration}, Brier Score~\cite{brier1950verification}, and AUROC~\cite{hanley1982meaning}.

\subsection{Factorial Perturbation Design}
\label{sec:perturbation}
\begin{figure*}[t]
\centering
\includegraphics[width=0.93\textwidth]{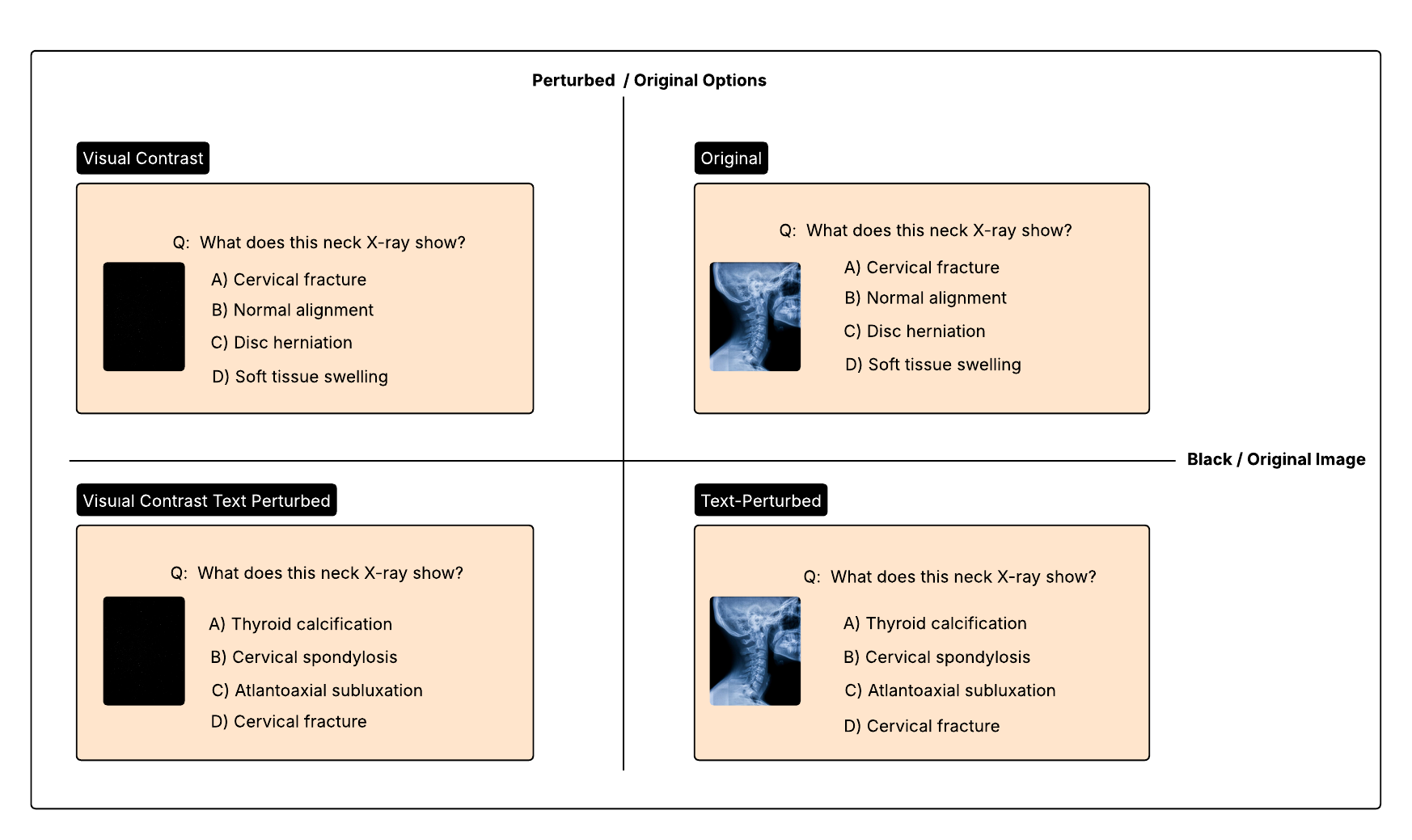}
\caption{The $2\times2$ factorial perturbation design illustrated on a cervical X-ray example from OmniMedVQA.}
\label{fig:perturbation_design}
\end{figure*}

In Medical VQA, the ground truth is by definition embedded in the image modality: answering correctly requires interpreting the visual evidence. Textual cues such as option co-occurrence patterns or positional biases often act as dataset-specific shortcuts rather than reflecting visually-grounded diagnostic reasoning. We therefore hypothesize that well-calibrated confidence should reflect the contribution of each evidence source to the prediction, and we design a perturbation scheme that teaches the model to ground its confidence in evidence utilization.

We cross two perturbation axes, image availability (original versus black image) and text integrity (original versus perturbed options), in a $2 \times 2$ factorial design that yields four conditions per question (see Figure~\ref{fig:perturbation_design}).
The black-image condition removes all visual evidence while preserving the input structure, probing the model's reliance on the visual modality rather than degrading it with classical perturbations such as Gaussian noise, which distort diagnostic features in medical images~\cite{liao2025visionamplified, avestimehr2025detecting}. We discuss the design rationale and provide qualitative examples of model behavior under the black-image condition in Section~\ref{sec:a_black_image} (Appendix). The perturbed-options condition shuffles the option order and replaces all distractors with options from other same-type questions (e.g., disease diagnosis from disease diagnosis), destroying memorized option patterns while keeping the correct answer and question unchanged.

For each sample under each condition, we generate $G = 10$ independent responses at temperature $T = 1.0$ and compute the fractional accuracy:
\begin{equation}
\hat{a}_j(i) = \frac{1}{G} \sum_{g=1}^{G} y_{i,j}^{(g)},
\label{eq:fractional_acc}
\end{equation}
where $y_{i,j}^{(g)} \in \{0, 1\}$ is the correctness of the $g$-th response for sample $i$ under condition $j$. These fractional estimates approximate the true correctness probability $p_i$ and serve as training targets for the Brier loss, whose optimum is $\hat{c}_i^* = p_i$, (see Section \ref{sec:a1_accuracy_estimation}).

\subsection{Composite Calibration Loss}

During training, the model autoregressively generates a sequence consisting of three segments: a reasoning trace, an answer, and a verbalized confidence token. Following ConfTuner's tokenized Brier score methodology~\cite{li2025conftuner}, the confidence token is drawn from a discrete integer scale determined by the model's available single-token integers and normalized to $[0, 1]$ by dividing by the scale maximum $N_A$. We distinguish two sets of token positions in the generated output: answer positions $\mathcal{T}$, spanning the reasoning and answer segments, and the confidence position, where the model produces a softmax distribution $p_\theta(v \mid i, j)$ over confidence tokens $v \in \{0, \ldots, N_A\}$. Our calibration losses operate directly on this softmax distribution rather than on a single decoded value, preserving the ordinal structure of the confidence scale through training.

Three calibration losses shape the model's expressed uncertainty: the Brier loss drives confidence toward the empirical correctness probability, the anchor loss prevents collapse toward extreme values, and the alignment loss ensures that confidence tracks evidence utilization across perturbation conditions. A separate KL divergence regularizer operates on the answer token positions to preserve the model's answering ability during confidence training.

\paragraph{Brier loss with fractional targets.} For sample $i$ under condition $j$:
\begin{equation}
\mathcal{L}_{\text{Brier}}(i, j)\!\! =\!\! \sum_{v=0}^{N_A} p_\theta(v \mid i, j) \left(\hat{a}_j(i) - v/N_A\right)^2
\label{eq:brier}
\end{equation}
This is the expected Brier score under the model's confidence distribution, driving probability mass toward the token whose value matches the fractional accuracy target.

\paragraph{Anchor loss.} Under the Brier loss, the optimal confidence for a sample with true correctness probability $p_i$ is $\hat{c}_i^* = p_i$ (Eq.~\ref{eq:brier_optimum}). For samples the model cannot reliably answer ($p_i \approx 0.5$), the optimum is the midpoint of the confidence scale, representing maximum uncertainty with respect to binary correctness. Under ideal optimization conditions, the Brier loss would induce this behavior, but in practice the model collapses toward extreme confidence values. The anchor loss addresses this by penalizing deviation from $c_{\text{mid}} = 0.5$, the point equidistant from both Brier targets ($y=0$ and $y=1$). This divides the scale into a negative side (0.0--0.4), where the model expresses belief that its answer is incorrect, and a positive side (0.6--1.0), where it expresses belief in correctness. The anchor acts as a conservative prior: the model starts at maximum uncertainty and must accumulate sufficient calibration signal from the Brier loss to move its confidence in either direction, preventing collapse toward extreme values without evidence:
\begin{equation}
\begin{split}
&\mathcal{L}_{\text{anchor}}(i, j) = \\ &\sum_{v=0}^{N_A} p_\theta(v \mid i, j) \left(v/N_A - c_{\text{mid}}\right)^2
\end{split}
\label{eq:anchor}
\end{equation}

\paragraph{Alignment loss.} The Brier and anchor losses operate on each condition independently. The alignment loss operates on the expected confidence
\begin{equation}
\hat{c}_j(i) = \sum_{v=0}^{N_A} p_\theta(v \mid i, j) \cdot v/N_A
\end{equation}
enforcing that confidence differences across conditions track accuracy differences for each sample:
\begin{equation}
\begin{split}
\mathcal{L}_{\text{align}}(i, j, k) = \left[ \right. & \left(\hat{c}_j(i) - \hat{c}_k(i)\right) \\
& \left. - \left(\hat{a}_j(i) - \hat{a}_k(i)\right) \right]^2
    \label{eq:align}
\end{split}
\end{equation}
computed over four same-axis condition pairs:
\begin{equation}
    \begin{split}
    \mathcal{P} = \{&(\text{V}, \text{VC}),\; (\text{V-TP}, \text{VC-TP}),\\
    &(\text{V}, \text{V-TP}),\; (\text{VC}, \text{VC-TP})\}.
    \end{split}
\label{eq:pairs}
\end{equation}
This teaches the model to ground its confidence in evidence utilization: the confidence change when an evidence source is removed should track the empirical accuracy change, reflecting the informational contribution of that source to the prediction.

\paragraph{Top-$k$ KL divergence regularizer.} Since confidence training modifies the model's shared representations through the LoRA adapters, it can also affect the answer distribution. We constrain this with a top-$k$ KL divergence on the answer token positions against a frozen copy of the base model:
\begin{equation}
    \begin{split}
    &\mathcal{L}_{\text{KL}}(i, j) = \\ 
    &\frac{1}{|\mathcal{T}|} \sum_{t \in \mathcal{T}} \sum_{v \in \mathcal{V}_k(t)} p_{\text{ref}}(v \mid t) \log \frac{p_{\text{ref}}(v \mid t)}{p_\theta(v \mid t)}
    \label{eq:kl}
    \end{split}
\end{equation}
where $\mathcal{T}$ is the set of generated token positions excluding the confidence token, $\mathcal{V}_k(t)$ contains the $k$ highest-probability tokens under the reference model at position $t$, $p_{\text{ref}}$ is the frozen base model distribution, and $p_\theta$ is the fine-tuned model distribution. Restricting the penalty to the top-$k$ tokens preserves the answer distribution over the candidates that determine the model's answer selection, while allowing representation changes in the remaining probability mass for confidence learning. Unlike reinforcement learning settings where KL regularization requires keeping a frozen reference model in memory and recomputing reference distributions at each training step, our setting uses fixed training data. This allows us to extract the reference distributions during data generation, which already requires a base model forward pass to produce the training pairs, and cache them offline. The KL term therefore adds no additional model copies or forward passes during training.

\paragraph{Complete objective.} We combine the four components as:
\begin{equation}
\begin{split}
\mathcal{L} = ~&\lambda\, \mathcal{L}_{\text{Brier}} + (1{-}\lambda)\, \mathcal{L}_{\text{anchor}} \\&+ \alpha\, \mathcal{L}_{\text{align}} + \beta\, \mathcal{L}_{\text{KL}}
\label{eq:total}
\end{split}
\end{equation}
where each term is averaged over all samples and four conditions. The coefficient $\lambda$ balances calibration against the conservative prior, $\alpha$ controls the strength of evidence-aware alignment, and $\beta$ controls answer preservation. We fine-tune the base model using LoRA adapters~\cite{hu2022lora}, targeting the query and value projections (3.2M trainable parameters, 0.075\% for MedGemma-4B-IT; 2.5M parameters, 0.030\% for Qwen2-VL-7B-Instruct).

%-----------------------------------------------------------------------------
% EXPERIMENTS AND RESULTS
%-----------------------------------------------------------------------------
\begin{table*}[!th]
\centering
\small
\begin{tabular}{llcccc}
\toprule
\textbf{Test Set} & \textbf{Method} & \textbf{ECE}$\downarrow$ & \textbf{Brier}$\downarrow$ & \textbf{AUROC}$\uparrow$ & \textbf{Acc.} \\
\midrule
& Base Model & 0.146 & 0.219 & 0.628 & 0.705 \\
& Top-K Sampling & 0.403$_{\pm.002}$ & 0.406$_{\pm.002}$ & 0.553$_{\pm.005}$ & 0.459$_{\pm.003}$ \\
\textbf{OmniMedVQA (ID)} & SteerConf & \textbf{0.050}$_{\pm.006}$ & 0.196$_{\pm.001}$ & 0.693$_{\pm.005}$ & 0.696$_{\pm.001}$ \\
& ConfTuner & \underline{0.074} & \underline{0.194} & \underline{0.708} & 0.693 \\
& Ours & 0.184 & \textbf{0.111}$^*$ & \textbf{0.926}$^*$ & 0.805 \\
\hline
& Base Model & 0.445 & 0.445 & 0.533 & 0.455 \\
& Top-K Sampling & 0.569$_{\pm.002}$ & 0.541$_{\pm.002}$ & 0.535$_{\pm.006}$ & 0.319$_{\pm.002}$ \\
\textbf{PMC-VQA (OOD)} & SteerConf & 0.302$_{\pm.003}$ & 0.336$_{\pm.002}$ & \underline{0.615}$_{\pm.004}$ & 0.462$_{\pm.002}$ \\
& ConfTuner & \underline{0.299} & \underline{0.329} & 0.583 & 0.436 \\
& Ours & \textbf{0.097}$^*$ & \textbf{0.243}$^*$ & \textbf{0.640}$^*$ & 0.463 \\
\hline
& Base Model & 0.656 & 0.617 & 0.486 & 0.238 \\
& Top-K Sampling & 0.686$_{\pm.003}$ & 0.628$_{\pm.002}$ & 0.471$_{\pm.006}$ & 0.184$_{\pm.003}$ \\
\textbf{MedXpertQA (OOD)} & SteerConf & \underline{0.487}$_{\pm.005}$ & \underline{0.448}$_{\pm.002}$ & 0.501$_{\pm.011}$ & 0.238$_{\pm.005}$ \\
& ConfTuner & 0.561 & 0.494 & \underline{0.507} & 0.218 \\
& Ours & \textbf{0.184}$^*$ & \textbf{0.257}$^*$ & \textbf{0.514} & 0.234 \\
\hline
& Base Model & 0.416 & 0.427 & 0.549 & 0.466 \\
& Top-K Sampling & 0.553 & 0.525 & 0.520 & 0.321 \\
\textbf{Average} & SteerConf & \underline{0.280} & \underline{0.327} & \underline{0.603} & 0.465 \\
& ConfTuner & 0.311 & 0.339 & 0.599 & 0.449 \\
& Ours & \textbf{0.155} & \textbf{0.204} & \textbf{0.693} & 0.501 \\
\bottomrule
\end{tabular}
\caption{MedGemma-4B-IT results for the OmniMedVQA-trained pathway. Best values in \textbf{bold}, second best \underline{underlined}. $\downarrow$: lower is better, $\uparrow$: higher is better. $^*$Statistically significant vs.\ second-best ($p<0.05$, paired bootstrap, $B$=10k).}
\label{tab:main_results}
\end{table*}

\begin{table*}[t]
\centering
\small
\begin{tabular}{llcccc}
\toprule
\textbf{Test Set} & \textbf{Method} & \textbf{ECE}$\downarrow$ & \textbf{Brier}$\downarrow$ & \textbf{AUROC}$\uparrow$ & \textbf{Acc.} \\
\midrule
& Base Model & 0.381 & 0.381 & 0.567 & 0.587 \\
& Top-K Sampling & 0.184$_{\pm.004}$ & 0.259$_{\pm.002}$ & 0.669$_{\pm.006}$ & 0.606$_{\pm.004}$ \\
\textbf{OmniMedVQA (ID)} & SteerConf & \underline{0.082}$_{\pm.004}$ & 0.222$_{\pm.001}$ & 0.705$_{\pm.004}$ & 0.562$_{\pm.004}$ \\
& ConfTuner & \textbf{0.068} & \underline{0.148} & \underline{0.860} & 0.674 \\
& Ours & 0.136 & \textbf{0.143} & \textbf{0.884}$^*$ & 0.660 \\
\hline
& Base Model & 0.504 & 0.503 & 0.517 & 0.488 \\
& Top-K Sampling & 0.377$_{\pm.006}$ & 0.381$_{\pm.004}$ & 0.617$_{\pm.009}$ & 0.440$_{\pm.006}$ \\
\textbf{PMC-VQA (OOD)} & SteerConf & \underline{0.199}$_{\pm.006}$ & \underline{0.264}$_{\pm.003}$ & \underline{0.692}$_{\pm.006}$ & 0.484$_{\pm.005}$ \\
& ConfTuner & 0.229 & 0.295 & 0.672 & 0.503 \\
& Ours & \textbf{0.058}$^*$ & \textbf{0.216}$^*$ & \textbf{0.703} & 0.490 \\
\hline
& Base Model & 0.790 & 0.786 & 0.501 & 0.198 \\
& Top-K Sampling & 0.569$_{\pm.003}$ & 0.491$_{\pm.002}$ & 0.491$_{\pm.005}$ & 0.206$_{\pm.002}$ \\
\textbf{MedXpertQA (OOD)} & SteerConf & \underline{0.351}$_{\pm.006}$ & \underline{0.315}$_{\pm.005}$ & \textbf{0.520}$_{\pm.017}$ & 0.196$_{\pm.007}$ \\
& ConfTuner & 0.360 & 0.369 & \underline{0.504} & 0.209 \\
& Ours & \textbf{0.314}$^*$ & \textbf{0.278}$^*$ & 0.487 & 0.194 \\
\hline
& Base Model & 0.558 & 0.557 & 0.528 & 0.424 \\
& Top-K Sampling & 0.377 & 0.377 & 0.592 & 0.417 \\
\textbf{Average} & SteerConf & \underline{0.211} & \underline{0.267} & 0.639 & 0.414 \\
& ConfTuner & 0.219 & 0.271 & \underline{0.679} & 0.462 \\
& Ours & \textbf{0.169} & \textbf{0.212} & \textbf{0.691} & 0.448 \\
\bottomrule
\end{tabular}
\caption{Qwen2-VL-7B-Instruct results for the OmniMedVQA-trained pathway. Best values in \textbf{bold}, second best \underline{underlined}. $\downarrow$: lower is better, $\uparrow$: higher is better. $^*$Statistically significant vs.\ second-best ($p<0.05$, paired bootstrap, $B$=10k).}
\label{tab:qwen2_results}
\end{table*}

\section{Experiments and Results}
\label{sec:results}
A calibration method that performs well at only one accuracy level may be exploiting the dataset mean accuracy rather than learning genuine calibration. To test this, we evaluate on three Medical VQA benchmarks spanning different difficulty regimes: OmniMedVQA~\cite{hu2024omnimedvqa} (high-accuracy regime, 59--70\% base accuracy depending on architecture), PMC-VQA~\cite{zhang2024pmcvqa} (mid-accuracy regime, 45--49\%), and MedXpertQA~\cite{zuo2025medxpertqa} (low-accuracy regime, 20--24\%). We compare against four baselines: the base model (zero-shot verbalized confidence), Top-K Sampling and SteerConf, which aggregate multiple stochastic passes, and ConfTuner, which fine-tunes verbalized confidence using a tokenized Brier score. We evaluate on two architectures: MedGemma-4B-IT~\cite{sellergren2025medgemma} and Qwen2-VL-7B-Instruct~\cite{wang2024qwen2vl}. Our method and ConfTuner produce a confidence estimate from a single forward pass; we use greedy decoding for deterministic evaluation. Top-K Sampling and SteerConf require multiple stochastic forward passes per sample; we report mean and standard deviation over 10 independent runs.

\paragraph{Loss Hyperparameters.}
Table~\ref{tab:hyperparams} reports the hyperparameter values for the final OmniMedVQA-trained configuration. The search was conducted using a GridSearch approach, in the ranges indicated in Table~\ref{tab:hyperparams}. The resolution of the grid search is $0.1$ for $\lambda$, $0.5$ for $\alpha$, and $1.0$ for $\beta$. We select hyperparameters on the validation set by filtering to the top quartile of accuracy, then choosing the configuration with the highest AUROC, as it depends only on rank ordering and is therefore not tied to the difficulty or accuracy level of the dataset.

For the PMC-VQA-trained pathway, the same grid search and selection criterion were applied within the hyperparameter family identified on the OmniMedVQA pathway. The selected configuration uses $\lambda{=}0.7$, $\alpha{=}1.5$, and $\beta{=}4.0$, with batch size reduced to 1 group (4 samples) and learning rate set to $1.4 \times 10^{-5}$ (Table~\ref{tab:training_config}).

\begin{table}[ht]
\centering
\resizebox{\columnwidth}{!}{
\begin{tabular}{l l l l}
\hline
\textbf{Hyperparameter} & \textbf{Symbol} & \textbf{Value} & \textbf{Explored range} \\
\hline
Brier--Anchor interpol. & $\lambda$ & 0.2 & $\{0.0,0.1,\ldots,1.0\}$ \\
Alignment weight & $\alpha$ & 2.0 & $\{0.5,1.0,\ldots,3.0\}$ \\
KL divergence weight & $\beta$ & 1.0 & $\{1.0,2.0,\ldots,5.0\}$ \\
Top-$k$ for KL & $k$ & 15 & $\{5,10,15,50\}$ \\
KL temperature & $T_{\text{KL}}$ & 4 & $\{1,2,\ldots,5\}$ \\
\hline
\end{tabular}
}

\caption{Loss hyperparameters for the final OmniMedVQA-trained configuration. The explored range indicates the values tested during hyperparameter search.}
\label{tab:hyperparams}
\end{table}

\begin{figure*}[ht]
\centering
\subfloat[Base model]{\includegraphics[width=0.32\textwidth]{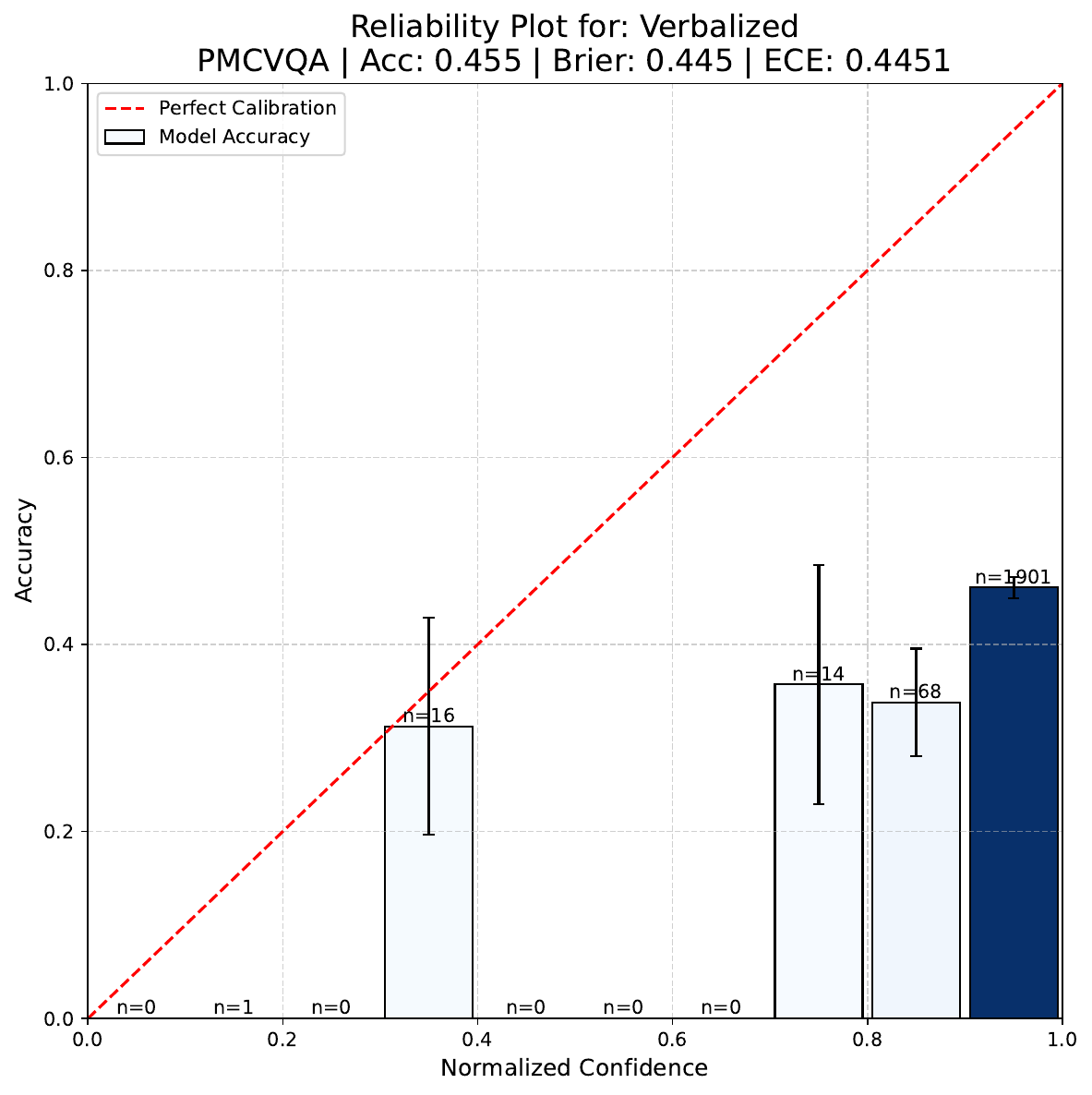}}
\hfill
\subfloat[SteerConf]{\includegraphics[width=0.32\textwidth]{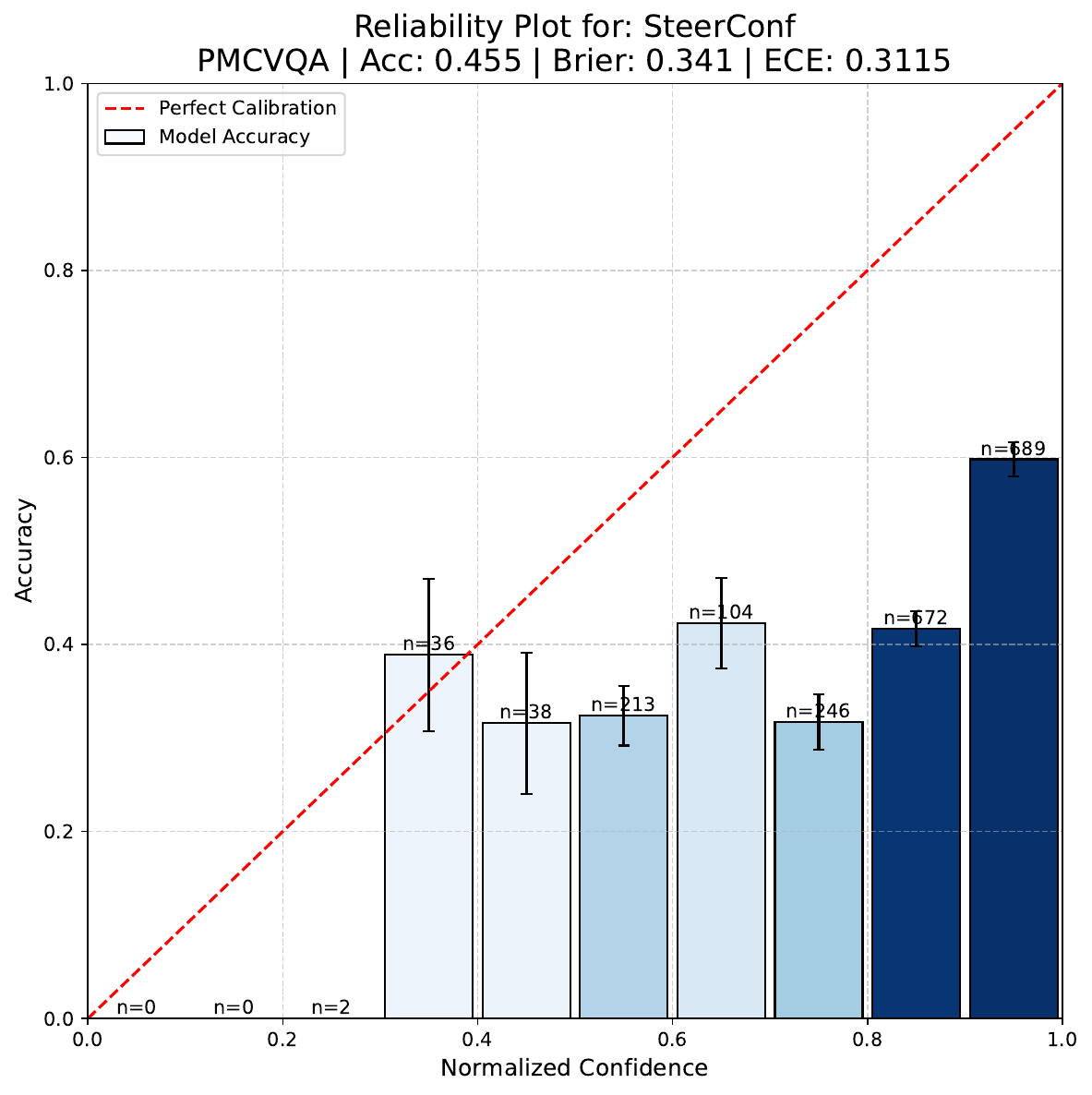}}
\hfill
\subfloat[Ours (PMC-trained)]{\includegraphics[width=0.32\textwidth]{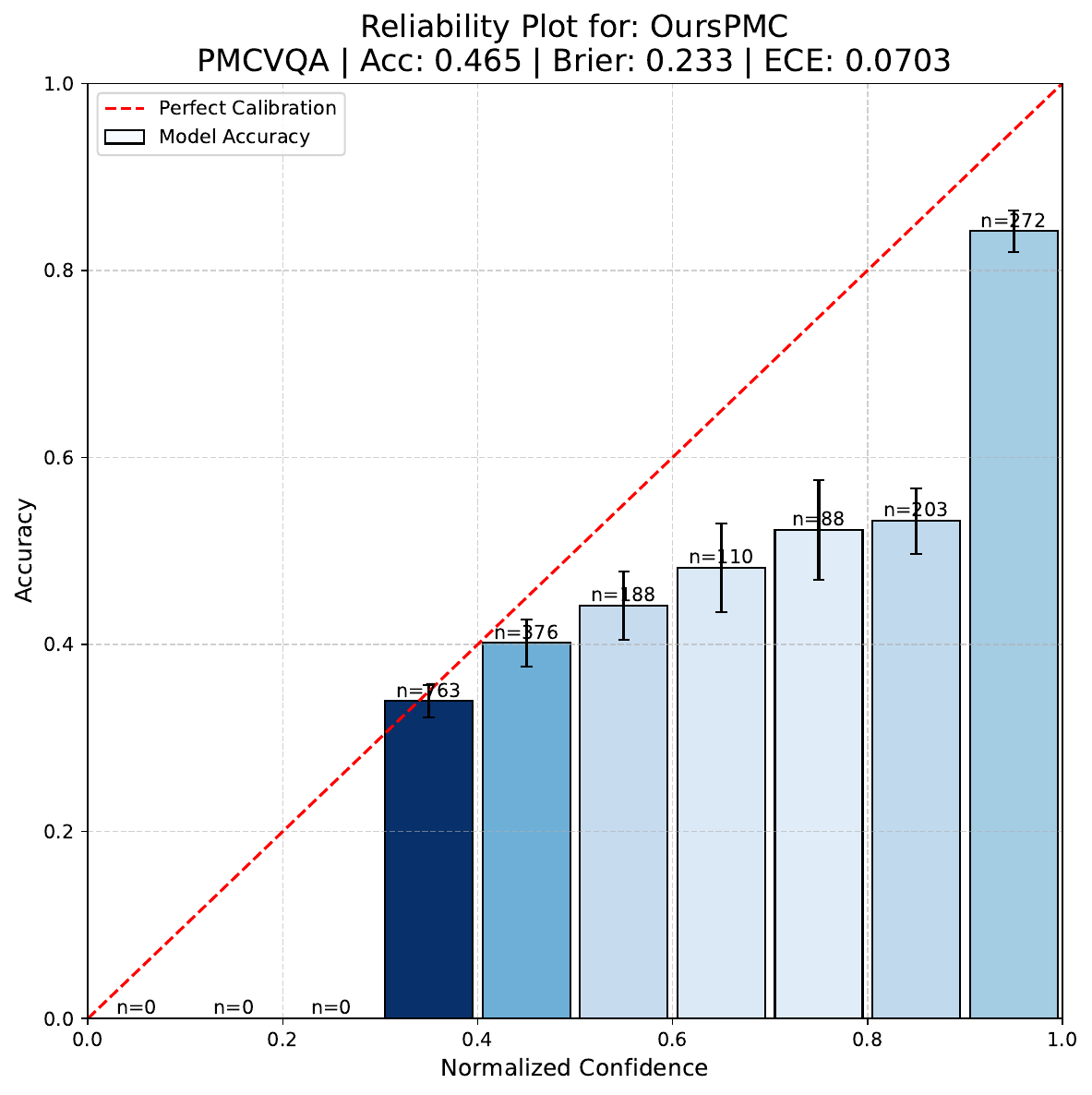}}
\caption{Reliability diagrams on PMC-VQA (MedGemma). Bar color indicates the number of samples in each bin (Sample count shown above each bin); error bars show the standard error of the bin accuracy.}
\label{fig:reliability_medgemma}
\end{figure*}

\paragraph{Main results.} Tables~\ref{tab:main_results} and~\ref{tab:qwen2_results} report results for MedGemma and Qwen respectively. Our method achieves the best average ECE, Brier Score, and AUROC under both architectures. The Brier Score, which jointly penalizes miscalibration and poor discrimination, is the strongest individual result: our method leads on every benchmark under both architectures, with reductions of 26--43\% relative to the next best method per benchmark on MedGemma and 3--18\% on Qwen. This consistency across high, mid, and low accuracy regimes indicates that the training objective captures a general relationship between confidence and correctness rather than exploiting a single benchmark's accuracy distribution. However, the improvement is not uniform across individual metrics: our method does not achieve the lowest ECE on OmniMedVQA or the highest AUROC on MedXpertQA for both MLLMs.

\paragraph{The calibration--discrimination trade-off.} On OmniMedVQA, baselines achieve substantially lower ECE (ConfTuner: 0.068 on Qwen, SteerConf: 0.050 on MedGemma) compared to our method (0.136 and 0.184 respectively). This low ECE comes at the cost of discrimination: on the same benchmark, our AUROC reaches 0.926 (MedGemma) and 0.884 (Qwen), while SteerConf remains at 0.693 and 0.705. The mechanism is confidence compression: methods that cluster predictions near the dataset mean achieve low ECE but cannot separate correct from incorrect predictions (Figures~\ref{fig:kde_compression}--\ref{fig:kde_cross_ours} and Table~\ref{tab:ovl}, Appendix~\ref{sec:a_kde}).

The difference is clearer across difficulty regimes. Baseline ECE degrades as accuracy decreases: on Qwen, ConfTuner's ECE rises from 0.068 to 0.229 to 0.360 across OmniMedVQA, PMC-VQA, and MedXpertQA; SteerConf follows the same trajectory (0.082 to 0.199 to 0.351). Our method produces the narrowest ECE spread under both architectures (0.097--0.184 on MedGemma, 0.058--0.314 on Qwen), indicating that calibration quality holds regardless of the accuracy regime. On MedXpertQA, no method achieves meaningful discrimination: AUROC ranges from 0.48 to 0.51 across all methods and both architectures. At approximately 20\% base accuracy, the signal for separating correct from incorrect predictions is insufficient for any verbalized confidence approach.

\begin{figure*}[t]
\centering
\subfloat[Base model]{\includegraphics[width=0.32\textwidth]{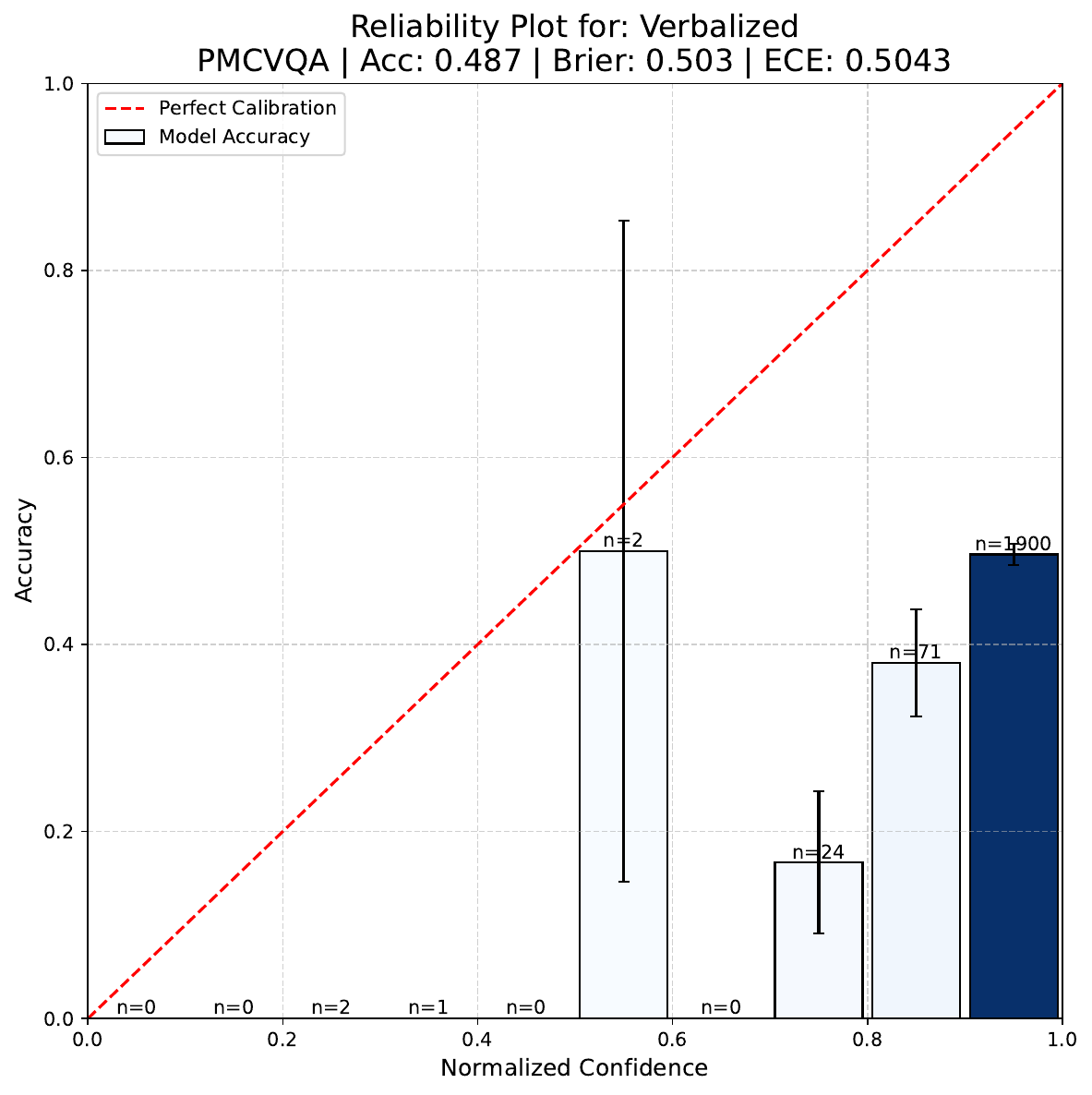}}
\hfill
\subfloat[SteerConf]{\includegraphics[width=0.32\textwidth]{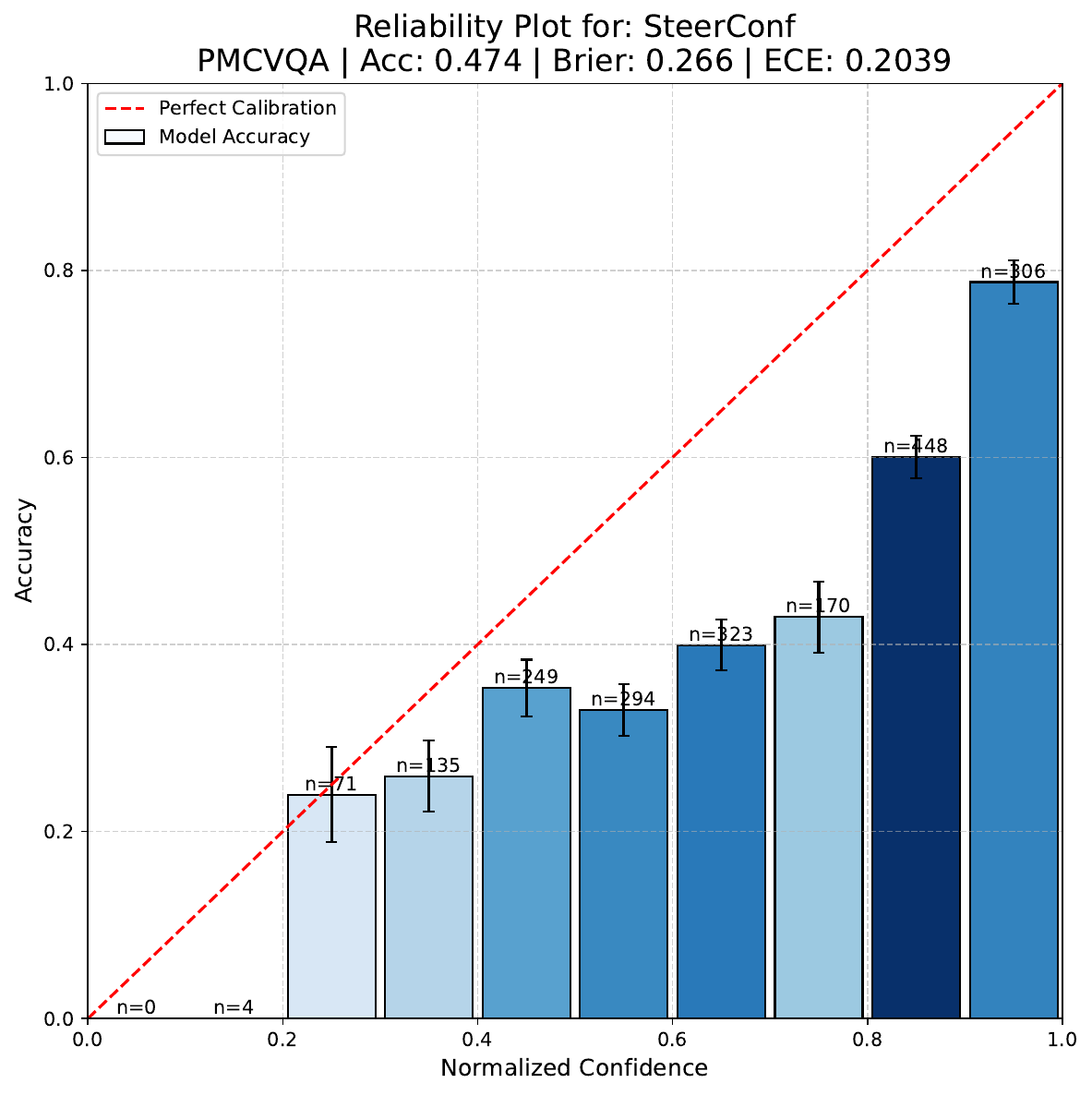}}
\hfill
\subfloat[Ours (OmniMed-trained)]{\includegraphics[width=0.32\textwidth]{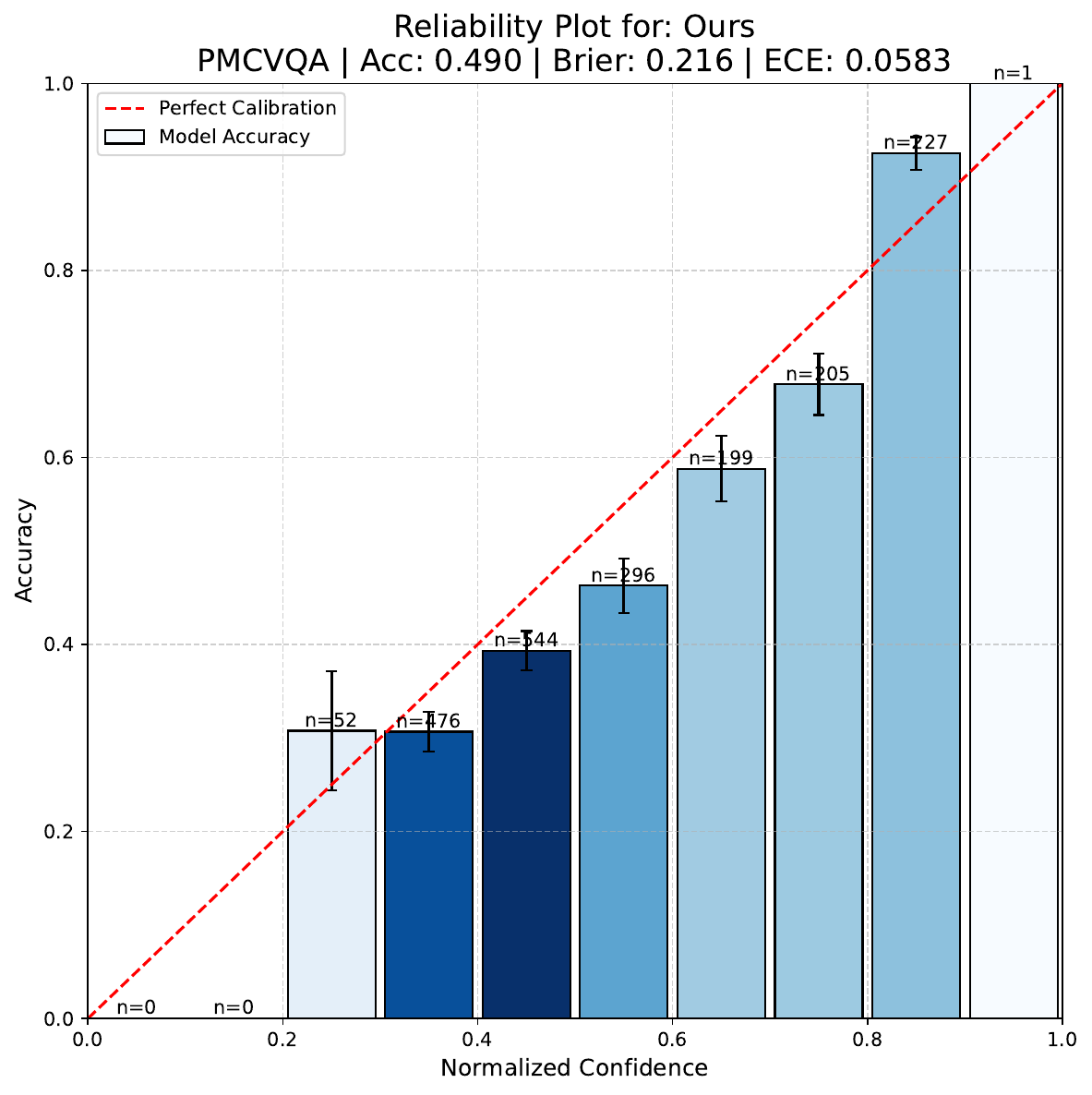}}
\caption{Reliability diagrams on PMC-VQA (Qwen2-VL). Bar color indicates the number of samples in each bin (Sample count shown above each bin); error bars show the standard error of the bin accuracy.}
\label{fig:reliability_qwen}
\end{figure*}

\paragraph{Accuracy preservation.} Our method preserves accuracy relative to the base model under both architectures: OOD accuracy changes are within $\pm 0.8$ percentage points, while ID accuracy increases by $+10.0$ (MedGemma) and $+7.3$ (Qwen) points on OmniMedVQA. The top-$k$ KL divergence regularizer anchors the fine-tuned distribution to a frozen copy of the base model, providing consistent accuracy preservation across both architectures. The ID accuracy gain reflects the operating point rather than a training objective; whether calibration gains persist under stricter accuracy constraints remains an open question for future work.

\paragraph{Cross-dataset transfer.} To test whether the training objective generalizes beyond the source distribution, we train a MedGemma model on PMC-VQA instead of OmniMedVQA and evaluate on all three benchmarks (Table~\ref{tab:transfer}; per-benchmark results in Table~\ref{tab:pmc_pathway_results}, Appendix). ConfTuner is re-trained on the same PMC-VQA split for a symmetric comparison.

\begin{table}[t]
\centering
\resizebox{\columnwidth}{!}{%
\begin{tabular}{lcccc}
\toprule
\textbf{Method} & \textbf{ECE}$\downarrow$ & \textbf{Brier}$\downarrow$ & \textbf{AUROC}$\uparrow$ & \textbf{Acc.} \\
\midrule
Base Model & 0.416 & 0.427 & 0.549 & 0.466 \\
Top-K Sampling & 0.553 & 0.525 & 0.520 & 0.321 \\
SteerConf & \underline{0.280} & \underline{0.327} & \underline{0.603} & 0.465 \\
ConfTuner & 0.437 & 0.445 & 0.563 & 0.459 \\
Ours (PMC-trained) & \textbf{0.123} & \textbf{0.226} & \textbf{0.624} & 0.461 \\
\bottomrule
\end{tabular}}
\caption{Cross-dataset transfer (MedGemma): average results across all three benchmarks for the PMC-VQA-trained pathway. Best in \textbf{bold}, second best \underline{underlined}.}
\label{tab:transfer}
\end{table}

The PMC-trained model achieves the best average ECE (0.123), Brier Score (0.226), and AUROC (0.624), outperforming all baselines on calibration and discrimination. The OmniMedVQA-trained pathway (Tables~\ref{tab:main_results},~\ref{tab:qwen2_results}) yields stronger overall results, but the PMC-trained pathway still outperforms all baselines on average, confirming that the training objective is not specific to a single source dataset. Figures~\ref{fig:reliability_medgemma} and~\ref{fig:reliability_qwen} show reliability diagrams on PMC-VQA for both architectures.

\paragraph{Component ablation.} Pairwise ablations on both architectures isolate the contribution of each loss component:
\begin{itemize}[leftmargin=*, nosep]
    \item \textbf{Without alignment} ($\alpha = 0$): Removing the alignment term causes AUROC to drop sharply (by 20.5\% on OmniMedVQA and 12.2\% on PMC-VQA) and Brier Score increases by 72.1\% on OmniMedVQA (Table~\ref{tab:ablation_noalign}). ECE improves on OmniMedVQA (9.8\%), indicating that confidence values become more conservative but less discriminative. Without the alignment term, confidence scores are well-calibrated but less informative.
    \item \textbf{Without KL} ($\beta = 0$): Removing the KL regularizer causes accuracy collapse on out-of-distribution data: $-15.0$ percentage points on PMC-VQA and $-11.3$ points on MedXpertQA, while ID accuracy decreases modestly ($-2.0$ points) (Table~\ref{tab:ablation_nokl}). Without the KL constraint, the model overfits to the confidence training signal at the expense of maintaining its question-answering capabilities.
\end{itemize}

On Qwen, the same component roles hold (Tables~\ref{tab:qwen_ablation_noalign},~\ref{tab:qwen_ablation_nokl}, Appendix): removing alignment preserves accuracy but drops AUROC from 0.884 to 0.835 on OmniMedVQA, and removing KL degrades accuracy by $-11.2$ points on OmniMedVQA and $-16.2$ on PMC-VQA. Removing KL additionally causes confidence format drift: approximately 22\% of outputs revert to alternative formats, producing confidence as ratios, on a 0--100 scale, or as linguistic terms instead of the expected integer scale (Table~\ref{tab:qwen_format}), while the full model maintains 100\% format compliance. The base model already follows the confidence format from pretraining; without the KL constraint, fine-tuning erodes this ability, showing that the regularizer preserves not only the answer distribution but also the model's existing instruction-following capabilities.

\paragraph{Practical implications.} To examine whether these calibration gains translate to deployment-relevant behavior, we analyze two simplified clinical scenarios in Appendix~\ref{sec:a_deployment}: an AI-assisted setting where the model recommends only when confident, and a clinician-assisted setting where low-confidence predictions are routed for human review. Our method adapts its recommendation volume to dataset difficulty and catches a larger share of errors across the full threshold range compared to all baselines.

%-----------------------------------------------------------------------------
% CONCLUSION
%-----------------------------------------------------------------------------
\section{Conclusions}
\label{sec:conclusions}

We presented a training-based framework for verbalized confidence calibration in Medical VQA, built on three contributions: a $2\times2$ factorial perturbation design that exposes how the model uses visual and textual evidence, a composite calibration loss that combines Brier-based calibration with contrastive alignment across perturbation conditions, and a top-$k$ KL divergence regularizer that preserves the answer distribution during confidence training.

Our results show that MLLMs can learn meaningful verbalized confidence through targeted fine-tuning. Under both MedGemma and Qwen2-VL, our method achieves the best average ECE, Brier Score, and AUROC across all three benchmarks, with calibration and discrimination improving simultaneously across three accuracy regimes. The learned confidence behavior generalizes under distribution shift: on MedGemma, both training pathways beat all baselines on average metrics, and consistent improvements on Qwen suggest that the gains are not architecture-specific. Accuracy is preserved within practical margins when the KL regularizer constrains the answer distribution. The ablation study shows that each loss component serves a distinct role: the alignment term drives discrimination, while the KL regularizer prevents forgetting and preserves the model's instruction-following capabilities, including confidence format compliance.

Two directions for future work emerge from these findings. Evaluating on larger model scales and extending to open-ended VQA would test whether the observed gains persist in more clinically realistic settings. On the methodological side, the anchor and alignment terms in our composite loss make explicit properties that the Brier loss encodes only implicitly. A calibration loss where these properties arise from the formulation itself, without requiring separate corrective terms, would be a more principled alternative.

\section*{Limitations}

Discrimination on MedXpertQA remains at chance level for all methods across both model architectures. This means verbalized confidence cannot meaningfully distinguish correct from incorrect predictions on questions requiring multi-step clinical reasoning, limiting the practical utility of calibrated confidence in the most challenging diagnostic scenarios. Both models in this work are small-scale (4B and 7B parameters). All experiments use multiple-choice VQA, where the model selects from a fixed set of options rather than generating free-form answers. This constrains the answer space in ways that simplify both prediction and calibration: correctness is binary and unambiguous. In open-ended clinical settings, where answers vary in granularity and partial correctness is common, calibrating verbalized confidence is a fundamentally harder problem that this work does not address.

\section*{Ethics Statement}

Uncalibrated Medical AI systems pose a direct safety risk: a model that expresses 90\% confidence when it is correct less than half the time actively misleads clinicians. Our method uses only publicly available benchmarks and does not make use of private clinical data. Medical deployment would therefore require further evaluation under real clinical conditions. Calibration quality may vary across patient populations, imaging modalities, and clinical contexts not represented in the evaluation datasets. This work evaluates calibration in a retrospective, multiple-choice setting; translating these gains to open-ended clinical tasks requires further validation. Finally and importantly, calibrated confidence estimation is intended to support clinical decision-making, but not to replace human judgment.

\bibliography{acl_latex}

@article{sellergren2025medgemma,
  title={MedGemma Technical Report},
  author={Sellergren, Andrew and Kazemzadeh, Sahar and Jaroensri, Tiam and Kiraly, Atilla and Traverse, Madeleine and Kohlberger, Timo and Xu, Shawn and Jamil, Fayaz and Hughes, C{\'\i}an and Lau, Charles and others},
  journal={arXiv preprint arXiv:2507.05201},
  year={2025},
  url={https://arxiv.org/abs/2507.05201}
}

@article{sheng2026nejm,
  title={Multimodal Large Language Models Challenge {NEJM} Image Challenge},
  author={Sheng, Chuyang and Shen, Shuo and Wang, Li and others},
  journal={Scientific Reports},
  year={2026},
  doi={10.1038/s41598-026-39201-3},
  publisher={Nature Publishing Group}
}

@article{nori2025sequential,
  title={Sequential Diagnosis with Language Models},
  author={Nori, Harsha and Daswani, Mayank and Kelly, Christopher and Lundberg, Scott and Ribeiro, Marco Tulio and Wilson, Marc and Liu, Xiaoxuan and Sounderajah, Viknesh and Carlson, Jonathan and Lungren, Matthew P. and Gross, Bay and Hames, Peter and Suleyman, Mustafa and King, Dominic and Horvitz, Eric},
  journal={arXiv preprint arXiv:2506.22405},
  year={2025}
}

@article{tian2023justask,
  title   = {Just Ask for Calibration: Strategies for Eliciting Calibrated Confidence Scores from Language Models Fine-Tuned with Human Feedback},
  author  = {Tian, Katherine and Mitchell, Eric and Zhou, Allan and Sharma, Archit and Rafailov, Rafael and Yao, Huaxiu and Finn, Chelsea and Manning, Christopher D.},
  journal = {arXiv preprint arXiv:2305.14975},
  year    = {2023}
}

@article{li2025conftuner,
  author        = {Li, Yibo and Xiong, Miao and Wu, Jiaying and Hooi, Bryan},
  title         = {ConfTuner: Training Large Language Models to Express Their Confidence Verbally},
  journal       = {arXiv preprint arXiv:2508.18847},
  year          = {2025},
  doi           = {10.48550/arXiv.2508.18847},
  url           = {https://arxiv.org/abs/2508.18847}
}

@article{xu2024sayself,
  title   = {{SaySelf}: Teaching {LLMs} to Express Confidence with Self-Reflective Rationales},
  author  = {Xu, Tianyang and Wu, Shujin and Diao, Shizhe and Liu, Xiaoze and Wang, Xingyao and Chen, Yangyi and Gao, Jing},
  journal = {arXiv preprint arXiv:2405.20974},
  year    = {2024}
}

@inproceedings{xuan2024vlmcalibration,
  title     = {Seeing is Believing, but How Much? A Comprehensive Analysis of Verbalized Calibration in Vision-Language Models},
  author    = {Xuan, Weihao and Zeng, Qingcheng and Qi, Heli and Wang, Junjue and Yokoya, Naoto},
  booktitle = {Proceedings of the 2025 Conference on Empirical Methods in Natural Language Processing},
  pages     = {1408--1450},
  year      = {2025},
  address   = {Suzhou, China},
  publisher = {Association for Computational Linguistics}
}

@article{kriz2025prompt4trust,
  title   = {{Prompt4Trust}: A Reinforcement Learning Prompt Augmentation Framework for Clinically-Aligned Confidence Calibration in Multimodal Large Language Models},
  author  = {Kriz, Anita and Janes, Elizabeth Laura and Shen, Xing and Arbel, Tal},
  journal = {arXiv preprint arXiv:2507.09279},
  year    = {2025}
}

@article{zhou2025steerconf,
  author        = {Zhou, Ziang and Jin, Tianyuan and Shi, Jieming and Li, Qing},
  title         = {SteerConf: Steering LLMs for Confidence Elicitation},
  journal       = {arXiv preprint arXiv:2503.02863},
  year          = {2025},
  url           = {https://arxiv.org/abs/2503.02863}
}

@inproceedings{xiong2024canllms,
  title   = {Can {LLMs} Express Their Uncertainty? An Empirical Evaluation of Confidence Elicitation in {LLMs}},
  author  = {Xiong, Miao and Hu, Zhiyuan and Lu, Xinyang and Li, Yifei and Fu, Jie and He, Junxian and Hooi, Bryan},
  booktitle = {International Conference on Learning Representations},
  year    = {2024}
}

@inproceedings{hu2022lora,
  title     = {LoRA: Low-Rank Adaptation of Large Language Models},
  author    = {Hu, Edward J. and Shen, Yelong and Wallis, Phillip and Allen-Zhu, Zeyuan and Li, Yuanzhi and Wang, Shean and Wang, Lu and Chen, Weizhu},
  booktitle = {ICLR},
  year      = {2022}
}

@inproceedings{bakman2024mars,
  title     = {{MARS}: Meaning-Aware Response Scoring for Uncertainty Estimation in Generative {LLMs}},
  author    = {Bakman, Yavuz Faruk and Yaldiz, Duygu Nur and Buyukates, Baturalp and Tao, Chenyang and Dimitriadis, Dimitrios and Avestimehr, Salman},
  booktitle = {Proceedings of the 62nd Annual Meeting of the Association for Computational Linguistics (ACL)},
  pages     = {7752--7767},
  year      = {2024}
}

@inproceedings{wang2023selfconsistency,
  title   = {Self-Consistency Improves Chain of Thought Reasoning in Language Models},
  author  = {Wang, Xuezhi and Wei, Jason and Schuurmans, Dale and Le, Quoc V. and Chi, Ed H. and Narang, Sharan and Chowdhery, Aakanksha and Zhou, Denny},
  booktitle = {International Conference on Learning Representations (ICLR)},
  year    = {2023}
}

@article{xiao2025eagle,
  title   = {Enhancing Uncertainty Estimation in {LLMs} with Expectation of Aggregated Internal Belief},
  author  = {Xiao, Zeguan and Dou, Diyang and Xiong, Boya and Chen, Yun and Chen, Guanhua},
  journal = {arXiv preprint arXiv:2509.01564},
  year    = {2025}
}

@article{yaldiz2024lars,
  title   = {Do Not Design, Learn: A Trainable Scoring Function for Uncertainty Estimation in Generative {LLMs}},
  author  = {Yaldiz, Duygu Nur and Bakman, Yavuz Faruk and Buyukates, Baturalp and Tao, Chenyang and Ramakrishna, Anil and Dimitriadis, Dimitrios and Zhao, Jieyu and Avestimehr, Salman},
  journal = {arXiv preprint arXiv:2406.11278},
  year    = {2024}
}

@article{stangel2025rewardingdoubt,
  title   = {Rewarding Doubt: A Reinforcement Learning Approach to Calibrated Confidence Expression of Large Language Models},
  author  = {Stangel, Paul and Bani-Harouni, David and Pellegrini, Chantal and {\"O}zsoy, Ege and Zaripova, Kamilia and Keicher, Matthias and Navab, Nassir},
  journal = {arXiv preprint arXiv:2503.02623},
  year    = {2025}
}

@article{zhang2025lovec,
  title   = {Reinforcement Learning for Better Verbalized Confidence in Long-Form Generation},
  author  = {Zhang, Caiqi and Zhu, Xiaochen and Li, Chengzu and Collier, Nigel and Vlachos, Andreas},
  journal = {arXiv preprint arXiv:2505.23912},
  year    = {2025}
}

@article{kadavath2022language,
  title   = {Language Models (Mostly) Know What They Know},
  author  = {Kadavath, Saurav and Conerly, Tom and Askell, Amanda and Henighan, Tom and Drain, Dawn and Perez, Ethan and Schiefer, Nicholas and Hatfield-Dodds, Zac and DasSarma, Nova and Tran-Johnson, Eli and Johnston, Scott and El-Showk, Sheer and Jones, Andy and Elhage, Nelson and Hume, Tristan and Chen, Anna and Bai, Yuntao and Bowman, Sam and Fort, Stanislav and Ganguli, Deep and Hernandez, Danny and Jacobson, Josh and Kernion, Jackson and Kravec, Shauna and Lovitt, Liane and Ndousse, Kamal and Olsson, Catherine and Ringer, Sam and Amodei, Dario and Brown, Tom and Clark, Jack and Joseph, Nicholas and Mann, Ben and McCandlish, Sam and Olah, Chris and Kaplan, Jared},
  journal = {arXiv preprint arXiv:2207.05221},
  year    = {2022}
}

@article{chaudhry2024finetuning,
  title   = {Finetuning Language Models to Emit Linguistic Expressions of Uncertainty},
  author  = {Chaudhry, Arslan and Thiagarajan, Sridhar and Gorur, Dilan},
  journal = {arXiv preprint arXiv:2409.12180},
  year    = {2024}
}

@inproceedings{hu2024omnimedvqa,
  title     = {OmniMedVQA: A New Large-Scale Comprehensive Evaluation Benchmark for Medical LVLM},
  author    = {Hu, Yutao and Li, Tianbin and Lu, Quanfeng and Shao, Wenqi and He, Junjun and Qiao, Yu and Luo, Ping},
  booktitle = {Proceedings of the IEEE/CVF Conference on Computer Vision and Pattern Recognition},
  year      = {2024},
  eprint    = {2402.09181},
  archivePrefix = {arXiv},
  primaryClass  = {cs.CV}
}

@article{zhang2024pmcvqa,
  title   = {PMC-VQA: Visual Instruction Tuning for Medical Visual Question Answering},
  author  = {Zhang, Xiaoman and Wu, Chaoyi and Zhao, Ziheng and Lin, Weixiong and Zhang, Ya and Wang, Yanfeng and Xie, Weidi},
  journal = {Communications Medicine},
  volume  = {4},
  pages   = {233},
  year    = {2024},
  publisher = {Nature Publishing Group}
}

@inproceedings{zuo2025medxpertqa,
  title     = {MedXpertQA: Benchmarking Expert-Level Medical Reasoning and Understanding},
  author    = {Zuo, Yuxin and Qu, Shang and Li, Yifei and Chen, Zhangren and Zhu, Xuekai and Hua, Ermo and Zhang, Kaiyan and Ding, Ning and Zhou, Bowen},
  booktitle = {ICML},
  year      = {2025}
}

@article{lin2022teaching,
  title   = {Teaching Models to Express Their Uncertainty in Words},
  author  = {Lin, Stephanie and Hilton, Jacob and Evans, Owain},
  journal = {TMLR},
  year    = {2022}
}

@article{han2024lepe,
  title   = {Enhancing Confidence Expression in Large Language Models Through Learning from Past Experience},
  author  = {Han, Haixia and Li, Tingyun and Chen, Shisong and Shi, Jie and Du, Chengyu and Xiao, Yanghua and Liang, Jiaqing and Lin, Xin},
  journal = {arXiv preprint arXiv:2404.10315},
  year    = {2024}
}

@article{wang2024qwen2vl,
  title={Qwen2-VL: Enhancing Vision-Language Model's Perception of the World at Any Resolution},
  author={Wang, Peng and Bai, Shuai and Tan, Sinan and Wang, Shijie and Fan, Zhihao and Bai, Jinze and Chen, Keqin and Liu, Xuejing and Wang, Jialin and Ge, Wenbin and Fan, Yang and Dang, Kai and Du, Mengfei and Ren, Xuancheng and Men, Rui and Liu, Dayiheng and Zhou, Chang and Zhou, Jingren and Lin, Junyang},
  journal={arXiv preprint arXiv:2409.12191},
  year={2024}
}

@article{yang2025recalibrate,
  title   = {{ReCalibrate}: {RL} for Uncertainty-Aware Reasoning in {LLMs}},
  author  = {Damani, Mehul and Puri, Isha and Slocum, Stewart and Shenfeld, Idan and Andreas, Jacob},
  year    = {2025},
  note    = {OpenReview: \url{https://openreview.net/forum?id=hiiCjfRhZI}}
}

@article{zhang2024vluncertainty,
  title   = {{VL-Uncertainty}: Detecting Hallucination in Large Vision-Language Model via Uncertainty Estimation},
  author  = {Zhang, Ruiyang and Zhang, Hu and Zheng, Zhedong},
  journal = {arXiv preprint arXiv:2411.11919},
  year    = {2024}
}

@inproceedings{malinin2021uncertainty,
  title     = {Uncertainty Estimation in Autoregressive Structured Prediction},
  author    = {Malinin, Andrey and Gales, Mark},
  booktitle = {International Conference on Learning Representations},
  year      = {2021}
}

@article{padhi2025harmony,
  title   = {{HARMONY}: Hidden Activation Representations and Model Output-Aware Uncertainty Estimation for Vision-Language Models},
  author  = {Padhi, Indu and Chen, Pin-Yu and Baldini, Ioana and Ramamurthy, Karthikeyan Natesan},
  journal = {arXiv preprint arXiv:2504.12345},
  year    = {2025}
}

@article{liu2024actcab,
  title   = {Enhancing Language Model Factuality via Activation-Based Confidence Calibration and Guided Decoding},
  author  = {Liu, Xin and Akter, Mu and Li, Yufei and Wu, Chien-Sheng and Xiong, Caiming},
  journal = {arXiv preprint arXiv:2412.02778},
  year    = {2024}
}

@inproceedings{avestimehr2025detecting,
  title   = {Detecting Unreliable Responses in Vision-Language Models via Visual Uncertainty},
  author  = {Avestimehr, Kiana and Aye, Emily and Fabian, Zalan and Mushtaq, Erum},
  booktitle = {ICLR 2025 Workshop on QUESTION},
  year    = {2025}
}

@inproceedings{kuhn2023semantic,
  title   = {Semantic Uncertainty: Linguistic Invariances for Uncertainty Estimation in Natural Language Generation},
  author  = {Kuhn, Lorenz and Gal, Yarin and Farquhar, Sebastian},
  booktitle = {International Conference on Learning Representations (ICLR)},
  year    = {2023}
}

@article{zhao2025csp,
  title   = {Object-Level Verbalized Confidence Calibration in Vision-Language Models via Semantic Perturbation},
  author  = {Zhao, Yunpu and Zhang, Rui and Xiao, Junbin and Hou, Ruibo and Guo, Jiaming and Zhang, Zihao and Hao, Yifan and Chen, Yunji},
  journal = {arXiv preprint arXiv:2504.08750},
  year    = {2025}
}

@InProceedings{liao2025visionamplified,
  author        = {Liao, Zehui and Hu, Shishuai and Zou, Ke and Fu, Huazhu and Zhen, Liangli and Xia, Yong},
  title         = {Vision-Amplified Semantic Entropy for Hallucination Detection in Medical Visual Question Answering},
  booktitle     = {Proceedings of Medical Image Computing and Computer Assisted Intervention -- MICCAI 2025},
  year          = {2025},
  publisher     = {Springer Nature Switzerland},
  volume        = {LNCS 15964},
  pages         = {669--679}
}

@article{hanley1982meaning,
  title={The meaning and use of the area under a receiver operating characteristic (ROC) curve},
  author={Hanley, James A and McNeil, Barbara J},
  journal={Radiology},
  volume={143},
  number={1},
  pages={29--36},
  year={1982},
  publisher={Radiological Society of North America}
}

@inproceedings{guo2017calibration,
  title     = {On Calibration of Modern Neural Networks},
  author    = {Guo, Chuan and Pleiss, Geoff and Sun, Yu and Weinberger, Kilian Q.},
  booktitle = {ICML},
  year      = {2017}
}

@article{brier1950verification,
  title   = {Verification of Forecasts Expressed in Terms of Probability},
  author  = {Brier, Glenn W.},
  journal = {Monthly Weather Review},
  volume  = {78},
  number  = {1},
  pages   = {1--3},
  year    = {1950}
}

@article{groot2024overconfidence,
  title   = {Overconfidence is Key: Verbalized Uncertainty Evaluation in Large Language and Vision-Language Models},
  author  = {Groot, Tobias and Valdenegro-Toro, Matias},
  journal = {arXiv preprint arXiv:2405.02917},
  year    = {2024}
}

@inproceedings{leng2024taming,
  title     = {Taming Overconfidence in {LLMs}: Reward Calibration in {RLHF}},
  author    = {Leng, Jixuan and Huang, Chengsong and Zhu, Banghua and Huang, Jiaxin},
  booktitle = {ICLR},
  year      = {2025}
}

@inproceedings{xiao2025restoring,
  title     = {Restoring Calibration for Aligned Large Language Models: A Calibration-Aware Fine-Tuning Approach},
  author    = {Xiao, Jiancong and Hou, Bojian and Wang, Zhanliang and Jin, Ruochen and Long, Qi and Su, Weijie J. and Shen, Li},
  booktitle = {ICML},
  year      = {2025}
}

\clearpage
\appendix

\section{Appendix}
\label{sec:appendix}

\subsection{Experimental Details}
\label{sec:a_experimental_details}

Table~\ref{tab:dataset_stats} reports the dataset splits used in our experiments. For OmniMedVQA, we construct a 20,000-sample pool from the open-access subset using quality-aware stratified sampling that preserves modality proportions and excludes low-quality source datasets (e.g., COVID-19-era collections with variable curation). From this pool, 4,000 training samples are randomly drawn; 3,000 test samples are drawn separately from the remaining open-access data using the same stratification. For PMC-VQA, we randomly subsample 4,000 training instances from the official train split and evaluate on a curated subset of 2,000 samples from the test split. MedXpertQA serves exclusively as an out-of-distribution benchmark, using 2,000 samples from its multimodal subset with no training.

\begin{table}[ht]
\centering
\small
\begin{tabular}{lcc}
\toprule
\textbf{Benchmark} & \textbf{Train} & \textbf{Test} \\
\midrule
OmniMedVQA & 4,000 & 3,000 \\
PMC-VQA & 4,000 & 2,000 \\
MedXpertQA & --- & 2,000 \\
\bottomrule
\end{tabular}
\caption{Dataset split sizes. MedXpertQA is used exclusively for out-of-distribution evaluation.}
\label{tab:dataset_stats}
\end{table}

Figure~\ref{fig:prompt_template} shows the prompt template used across all methods and benchmarks. The model receives the medical image, the question, and four answer options, and is asked to produce a rationale, an answer, and a verbalized confidence score on a scale of 1 to 10. For SteerConf, a steering instruction is inserted at the designated position; for all other methods, this field is left empty. Top-K Sampling uses a separate template that requests the model's two best guesses with individual confidence scores.

\begin{figure}[t]
\centering
\fbox{\parbox{0.93\columnwidth}{\small\ttfamily
\{question\}\\[2pt]
Choose exactly one answer from the four options below.\\
Options:\\
- A: \{option\_a\} \quad - B: \{option\_b\}\\
- C: \{option\_c\} \quad - D: \{option\_d\}\\[2pt]
Based on the image, provide the most likely medical finding or answer concisely by following the output format below.\\
\{STEER\_INSTRUCTION\}\\
Output format:\\
Rationale: <1--2 concise sentences>\\
Answer: <LETTER> --- <EXACT OPTION TEXT>\\
After the answer please provide your confidence on your answer in the following format on a scale of 1 to 10:\\
Confidence: <1-10>
}}
\caption{Prompt template used for the base model, our method, SteerConf, and ConfTuner. Top-K Sampling uses a variant that requests the model's two best guesses with individual confidence scores. Qwen2-VL uses the same prompt content with model-specific chat formatting.}
\label{fig:prompt_template}
\end{figure}

ConfTuner~\cite{li2025conftuner} was originally designed for text-only LLMs. We reimplemented it for multimodal models following the original code repository, training on the same 4,000-sample OmniMedVQA split without the perturbation augmentation used by our method. SteerConf~\cite{zhou2025steerconf} was implemented using the prompts and sampling parameters from the original code repository ($T{=}0.7$, 5 inference passes per sample). Top-K Sampling follows~\cite{xiong2024canllms}: the model produces its two best guesses with individual confidence scores ($k{=}2$), sampled at $T{=}0.7$ over 5 passes, and the final confidence is the average across responses. The alternative pair-rank aggregation proposed in the same work performed worse in our setting, so we report the average-confidence variant.

\subsection{Training Configuration}
\label{sec:a_training_config}

Table~\ref{tab:training_config} reports the hyperparameters for all training configurations. We use the AdamW optimizer with no learning rate scheduler and fine-tune with LoRA adapters~\cite{hu2022lora}. For MedGemma-4B-IT this yields 3.2M trainable parameters (0.075\% of 4.3B); for Qwen2-VL-7B-Instruct, 2.5M parameters (0.030\% of 8.3B) due to grouped-query attention reducing the value projection dimensions. All configurations train for 3 epochs; we select epoch 2 for both MedGemma pathways and epoch 3 for Qwen based on calibration metrics on the in-distribution validation set. Training runs use an NVIDIA RTX Pro 6000 (Blackwell) GPU, with each epoch taking approximately one hour.

\begin{table}[t]
\centering
\small
\resizebox{\columnwidth}{!}{%
\begin{tabular}{lccc}
\toprule
\textbf{Hyperparameter} & \textbf{MedGemma Omni} & \textbf{MedGemma PMC} & \textbf{Qwen Omni} \\
\midrule
LoRA rank $r$ & 8 & 8 & 8 \\
LoRA scaling $\alpha_{\text{LoRA}}$ & 32 & 32 & 32 \\
LoRA dropout & 0.05 & 0.05 & 0.05 \\
Learning rate & $2.1 \times 10^{-5}$ & $1.4 \times 10^{-5}$ & $1.4 \times 10^{-5}$ \\
Batch size & 2 groups (8) & 1 group (4) & 1 group (4) \\
Brier--Anchor $\lambda$ & 0.2 & 0.7 & 0.6 \\
Alignment $\alpha$ & 2.0 & 1.5 & 2.0 \\
KL weight $\beta$ & 1.0 & 4.0 & 1.0 \\
Top-$k$ for KL & 15 & 15 & 15 \\
Selected epoch & 2 & 2 & 3 \\
\bottomrule
\end{tabular}}
\caption{Training hyperparameters for all configurations. Each group consists of one sample and its three perturbation variants.}
\label{tab:training_config}
\end{table}

\subsection{Accuracy Estimation}
\label{sec:a1_accuracy_estimation}
In the multiple-choice setting with $K$ options, a single response yields a binary correctness label $y_i \in \{0, 1\}$. A random guesser achieves expected accuracy $1/K$, so a single correct response does not reliably distinguish genuine knowledge from a random guess. For instance, a model that has no understanding of a particular question and selects uniformly among $K = 4$ options still answers correctly 25\% of the time. To obtain a reliable estimate of whether the model actually knows the answer for a given sample, we generate $G$ independent responses for each sample under each condition and compute the fraction of correct responses:

\begin{equation}
\hat{a}_j(i) = \frac{1}{G} \sum_{g=1}^{G} y_{i,j}^{(g)}
\label{eq:per_sample_acc}
\end{equation}

where $y_{i,j}^{(g)} \in \{0, 1\}$ denotes the correctness of the $g$-th response for sample $i$ under condition $j \in \{\text{V}, \text{VC}, \text{V-TP}, \text{VC-TP}\}$. The resulting estimate $\hat{a}_j(i) \in \{0, \tfrac{1}{G}, \ldots, 1\}$ provides a fractional accuracy that estimates the model's true correctness probability $p_{i,j} = P(y_{i,j} = 1)$ under condition $j$. Since the $G$ responses are independent draws at temperature $T = 1.0$, $\hat{a}_j(i)$ is an unbiased estimator with variance $p_{i,j}(1 - p_{i,j}) / G$. We set $G = 10$, which balances variance reduction against computational cost.

This fractional estimate is the correct training target for the Brier loss. Let $p_i$ denote the model's true correctness probability for sample $i$. The expected Brier loss over the stochastic binary label is:

\begin{equation}
\mathbb{E}_{y_i}\!\left[(y_i - \hat{c}_i)^2\right] = p_i\,(1 - \hat{c}_i)^2 + (1 - p_i)\,\hat{c}_i^2
\label{eq:expected_brier}
\end{equation}

Differentiating with respect to the predicted confidence $\hat{c}_i$ and setting the result to zero gives the optimal prediction:

\begin{equation}
    \begin{split}
    \frac{\partial}{\partial \hat{c}_i}\,\mathbb{E}\!\left[(y_i - \hat{c}_i)^2\right] = -2p_i + 2\hat{c}_i = 0 \quad \\ \Longrightarrow \quad \hat{c}_i^* = p_i
    \label{eq:brier_optimum}
    \end{split}
\end{equation}

The Brier loss is therefore minimized when the predicted confidence equals the true correctness probability. Our fractional estimate $\hat{a}(i)$ directly approximates $p_i$, providing the model with a training target that is close to this theoretical optimum.

Given that $G$ inference runs are already necessary to resolve guessing noise, training on $G$ binary labels is less efficient than training on the single fractional label. Both approaches converge to the same optimum, as the following gradient identity shows:

\begin{equation}
    \begin{split}
    \nabla_{\hat{c}_i} \frac{1}{G}\sum_{g=1}^{G} (&y_i^{(g)} - \hat{c}_i)^2 \;=\; -2\!\left(\hat{a}(i) - \hat{c}_i\right) \;\\&=\; \nabla_{\hat{c}_i}\,(\hat{a}(i) - \hat{c}_i)^2
    \label{eq:gradient_identity}
    \end{split}
\end{equation}

The equality holds by linearity of the gradient and the definition of $\hat{a}(i)$. Averaging the Brier loss gradient over $G$ binary labels produces exactly the same update direction as computing the gradient from a single Brier loss with the fractional target $\hat{a}(i)$. The two formulations are therefore equivalent in their optimization signal, but the fractional formulation requires only one backward pass instead of $G$. Since both approaches already require $G$ forward-only inference passes to obtain the labels, training on the fractional estimate saves $G - 1$ backward passes, each of which is more expensive than an inference pass due to gradient computation and backpropagation. ConfTuner~\cite{li2025conftuner}, by contrast, trains on a single binary label from one inference run and does not have access to a fractional estimate of $p_i$.

\subsection{PMC-VQA-Trained Pathway: Per-Benchmark Results}
\label{sec:a2_pmc_fine_grained}

\begin{table}[]
\centering
\resizebox{\columnwidth}{!}{%
\begin{tabular}{llcccc}
\toprule
\textbf{Test Set} & \textbf{Method} & \textbf{ECE} $\downarrow$ & \textbf{Brier} $\downarrow$ & \textbf{AUROC} $\uparrow$ & \textbf{Acc.} \\
\midrule
& Base Model         & 0.146 & 0.219 & 0.628 & 0.705 \\
& Top-K Sampling     & 0.403$_{\pm.002}$ & 0.406$_{\pm.002}$ & 0.553$_{\pm.005}$ & 0.459$_{\pm.003}$ \\
\textbf{OmniMedVQA (OOD)} & SteerConf & \textbf{0.050}$_{\pm.006}$ & \textbf{0.196}$_{\pm.001}$ & \underline{0.693}$_{\pm.005}$ & 0.696$_{\pm.001}$ \\
& ConfTuner          & 0.166 & 0.231 & 0.633 & 0.691 \\
& Ours (PMC-trained) & \underline{0.095} & \textbf{0.196} & \textbf{0.728}$^*$ & 0.688 \\
\hline
& Base Model         & 0.445 & 0.445 & 0.533 & 0.455 \\
& Top-K Sampling     & 0.569$_{\pm.003}$ & 0.541$_{\pm.002}$ & 0.535$_{\pm.006}$ & 0.319$_{\pm.002}$ \\
\textbf{PMC-VQA (ID)} & SteerConf & \underline{0.302}$_{\pm.003}$ & \underline{0.336}$_{\pm.002}$ & \underline{0.615}$_{\pm.004}$ & 0.462$_{\pm.002}$ \\
& ConfTuner          & 0.468 & 0.462 & 0.578 & 0.459 \\
& Ours (PMC-trained) & \textbf{0.070}$^*$ & \textbf{0.233}$^*$ & \textbf{0.660}$^*$ & 0.465 \\
\hline
& Base Model         & 0.656 & 0.617 & \underline{0.486} & 0.238 \\
& Top-K Sampling     & 0.686$_{\pm.003}$ & 0.628$_{\pm.002}$ & 0.471$_{\pm.006}$ & 0.184$_{\pm.003}$ \\
\textbf{MedXpertQA (OOD)} & SteerConf & \underline{0.487}$_{\pm.005}$ & \underline{0.448}$_{\pm.002}$ & \textbf{0.501}$_{\pm.011}$ & 0.238$_{\pm.005}$ \\
& ConfTuner          & 0.678 & 0.641 & 0.478 & 0.226 \\
& Ours (PMC-trained) & \textbf{0.205}$^*$ & \textbf{0.248}$^*$ & 0.483 & 0.229 \\
\midrule
& Base Model         & 0.416 & 0.427 & 0.549 & 0.466 \\
& Top-K Sampling     & 0.553 & 0.525 & 0.520 & 0.321 \\
\textbf{Average} & SteerConf & \underline{0.280} & \underline{0.327} & \underline{0.603} & 0.465 \\
& ConfTuner          & 0.437 & 0.445 & 0.563 & 0.459 \\
& Ours (PMC-trained) & \textbf{0.123} & \textbf{0.226} & \textbf{0.624} & 0.461 \\
\bottomrule
\end{tabular}%
}
\caption{Per-benchmark cross-dataset transfer results for the PMC-VQA-trained pathway. Best values in \textbf{bold}, second best \underline{underlined}. $^*$Statistically significant vs.\ second-best ($p<0.05$, paired bootstrap, $B$=10k).}
\label{tab:pmc_pathway_results}
\end{table}

Table~\ref{tab:pmc_pathway_results} reports the per-benchmark results for the PMC-VQA-trained pathway summarized in Table~\ref{tab:transfer}.

\subsection{Per-Benchmark Ablation Results}
\label{sec:a_ablation}

\begin{table*}[]
\centering
\resizebox{\textwidth}{!}{%
\begin{tabular}{lcccccccccccc}
\toprule
\textbf{Test Set} & \multicolumn{3}{c}{\textbf{ECE} $\downarrow$} & \multicolumn{3}{c}{\textbf{Brier} $\downarrow$} & \multicolumn{3}{c}{\textbf{AUROC} $\uparrow$} & \multicolumn{3}{c}{\textbf{Acc.}} \\
\cmidrule(lr){2-4}\cmidrule(lr){5-7}\cmidrule(lr){8-10}\cmidrule(lr){11-13}
& Full & No Align & $\Delta$ & Full & No Align & $\Delta$ & Full & No Align & $\Delta$ & Full & No Align & $\Delta$ \\
\midrule
\textbf{OmniMedVQA (ID)} & 0.184 & 0.166 & 9.8\% $\uparrow$ & 0.111 & 0.191 & 72.1\% $\downarrow$ & 0.926 & 0.736 & 20.5\% $\downarrow$ & 0.805 & 0.719 & 10.7\% $\downarrow$ \\
\textbf{PMC-VQA (OOD)} & 0.097 & 0.154 & 58.8\% $\downarrow$ & 0.243 & 0.267 & 9.9\% $\downarrow$ & 0.640 & 0.562 & 12.2\% $\downarrow$ & 0.463 & 0.465 & 0.4\% $\uparrow$ \\
\textbf{MedXpertQA (OOD)} & 0.184 & 0.379 & 106.0\% $\downarrow$ & 0.257 & 0.327 & 27.2\% $\downarrow$ & 0.514 & 0.504 & 1.9\% $\downarrow$ & 0.234 & 0.234 & 0.0\% \\
\bottomrule
\end{tabular}%
}
\caption{MedGemma pairwise ablation: full method versus no alignment ($\alpha=0$). $\uparrow$ = performance improved, $\downarrow$ = performance degraded.}
\label{tab:ablation_noalign}
\end{table*}

\begin{table*}[]
\centering
\resizebox{\textwidth}{!}{%
\begin{tabular}{lcccccccccccc}
\toprule
\textbf{Test Set} & \multicolumn{3}{c}{\textbf{ECE} $\downarrow$} & \multicolumn{3}{c}{\textbf{Brier} $\downarrow$} & \multicolumn{3}{c}{\textbf{AUROC} $\uparrow$} & \multicolumn{3}{c}{\textbf{Acc.}} \\
\cmidrule(lr){2-4}\cmidrule(lr){5-7}\cmidrule(lr){8-10}\cmidrule(lr){11-13}
& Full & No KL & $\Delta$ & Full & No KL & $\Delta$ & Full & No KL & $\Delta$ & Full & No KL & $\Delta$ \\
\midrule
\textbf{OmniMedVQA (ID)} & 0.184 & 0.141 & 23.4\% $\uparrow$ & 0.111 & 0.125 & 12.6\% $\downarrow$ & 0.926 & 0.877 & 5.3\% $\downarrow$ & 0.805 & 0.785 & 2.5\% $\downarrow$ \\
\textbf{PMC-VQA (OOD)} & 0.097 & 0.153 & 57.7\% $\downarrow$ & 0.243 & 0.213 & 12.3\% $\uparrow$ & 0.640 & 0.704 & 10.0\% $\uparrow$ & 0.463 & 0.313 & 32.4\% $\downarrow$ \\
\textbf{MedXpertQA (OOD)} & 0.184 & 0.390 & 112.0\% $\downarrow$ & 0.257 & 0.283 & 10.1\% $\downarrow$ & 0.514 & 0.613 & 19.3\% $\uparrow$ & 0.234 & 0.121 & 48.3\% $\downarrow$ \\
\bottomrule
\end{tabular}%
}
\caption{MedGemma pairwise ablation: full method versus no KL ($\beta=0$). $\uparrow$ = performance improved, $\downarrow$ = performance degraded.}
\label{tab:ablation_nokl}
\end{table*}

Tables~\ref{tab:ablation_noalign} and~\ref{tab:ablation_nokl} report the per-benchmark ablation results summarized in Section~\ref{sec:results}. The $\Delta$ column reports the relative change from Full to Ablation; arrows indicate the direction of performance change ($\uparrow$ = improved, $\downarrow$ = degraded).

\begin{table}[h]
\centering
\resizebox{\columnwidth}{!}{%
\begin{tabular}{lccccccc}
\toprule
\textbf{Model} & \textbf{0--9} & \textbf{10--100} & \textbf{Frac.} & \textbf{Words} & \textbf{Pct} & \textbf{Other} & \textbf{Parse} \\
\midrule
Full & 3000 & 0 & 0 & 0 & 0 & 0 & 100\% \\
No-KL & 2329 & 106 & 521 & 7 & 9 & 28 & 77.6\% \\
\bottomrule
\end{tabular}%
}
\caption{Confidence format compliance on OmniMedVQA (Qwen2-VL, $N{=}3000$). Columns show the number of outputs in each format category. Parse = percentage producing valid 0--9 integers.}
\label{tab:qwen_format}
\end{table}

\begin{table*}[h]
\centering
\resizebox{\textwidth}{!}{%
\begin{tabular}{lcccccccccccc}
\toprule
\textbf{Test Set} & \multicolumn{3}{c}{\textbf{ECE} $\downarrow$} & \multicolumn{3}{c}{\textbf{Brier} $\downarrow$} & \multicolumn{3}{c}{\textbf{AUROC} $\uparrow$} & \multicolumn{3}{c}{\textbf{Acc.}} \\
\cmidrule(lr){2-4}\cmidrule(lr){5-7}\cmidrule(lr){8-10}\cmidrule(lr){11-13}
& Full & No Align & $\Delta$ & Full & No Align & $\Delta$ & Full & No Align & $\Delta$ & Full & No Align & $\Delta$ \\
\midrule
\textbf{OmniMedVQA (ID)} & 0.136 & 0.200 & 47.1\% $\downarrow$ & 0.143 & 0.196 & 37.1\% $\downarrow$ & 0.884 & 0.835 & 5.5\% $\downarrow$ & 0.660 & 0.611 & 7.4\% $\downarrow$ \\
\textbf{PMC-VQA (OOD)} & 0.058 & 0.036 & 37.9\% $\uparrow$ & 0.216 & 0.224 & 3.7\% $\downarrow$ & 0.703 & 0.671 & 4.6\% $\downarrow$ & 0.490 & 0.481 & 1.8\% $\downarrow$ \\
\textbf{MedXpertQA (OOD)} & 0.314 & 0.217 & 30.9\% $\uparrow$ & 0.278 & 0.211 & 24.1\% $\uparrow$ & 0.487 & 0.497 & 2.1\% $\uparrow$ & 0.194 & 0.194 & 0.0\% \\
\bottomrule
\end{tabular}%
}
\caption{Qwen2-VL pairwise ablation: full method versus no alignment ($\alpha=0$). Unparseable confidence outputs assigned midpoint value. $\uparrow$ = improved, $\downarrow$ = degraded.}
\label{tab:qwen_ablation_noalign}
\end{table*}

\begin{table*}[h]
\centering
\resizebox{\textwidth}{!}{%
\begin{tabular}{lcccccccccccc}
\toprule
\textbf{Test Set} & \multicolumn{3}{c}{\textbf{ECE} $\downarrow$} & \multicolumn{3}{c}{\textbf{Brier} $\downarrow$} & \multicolumn{3}{c}{\textbf{AUROC} $\uparrow$} & \multicolumn{3}{c}{\textbf{Acc.}} \\
\cmidrule(lr){2-4}\cmidrule(lr){5-7}\cmidrule(lr){8-10}\cmidrule(lr){11-13}
& Full & No KL & $\Delta$ & Full & No KL & $\Delta$ & Full & No KL & $\Delta$ & Full & No KL & $\Delta$ \\
\midrule
\textbf{OmniMedVQA (ID)} & 0.136 & 0.066 & 51.5\% $\uparrow$ & 0.143 & 0.231 & 61.5\% $\downarrow$ & 0.884 & 0.656 & 25.8\% $\downarrow$ & 0.660 & 0.548 & 17.0\% $\downarrow$ \\
\textbf{PMC-VQA (OOD)} & 0.058 & 0.193 & 232.8\% $\downarrow$ & 0.216 & 0.249 & 15.3\% $\downarrow$ & 0.703 & 0.628 & 10.7\% $\downarrow$ & 0.490 & 0.328 & 33.1\% $\downarrow$ \\
\textbf{MedXpertQA (OOD)} & 0.314 & 0.349 & 11.1\% $\downarrow$ & 0.278 & 0.288 & 3.6\% $\downarrow$ & 0.487 & 0.517 & 6.2\% $\uparrow$ & 0.194 & 0.182 & 6.2\% $\downarrow$ \\
\bottomrule
\end{tabular}%
}
\caption{Qwen2-VL pairwise ablation: full method versus no KL ($\beta=0$). Unparseable confidence outputs assigned midpoint value. $\uparrow$ = improved, $\downarrow$ = degraded.}
\label{tab:qwen_ablation_nokl}
\end{table*}

Tables~\ref{tab:qwen_ablation_noalign} and~\ref{tab:qwen_ablation_nokl} report the Qwen2-VL ablation results discussed in Section~\ref{sec:results}. Removing KL causes confidence format drift: the model reverts to producing confidence as ratios, on a 0--100 scale, or as linguistic terms. Table~\ref{tab:qwen_format} reports the format compliance breakdown on OmniMedVQA. Unparseable samples in both ablation tables are assigned the midpoint value (maximum uncertainty).

\subsection{Confidence Distribution Analysis}
\label{sec:a_kde}

\begin{figure*}[t]
\centering
\includegraphics[width=\textwidth]{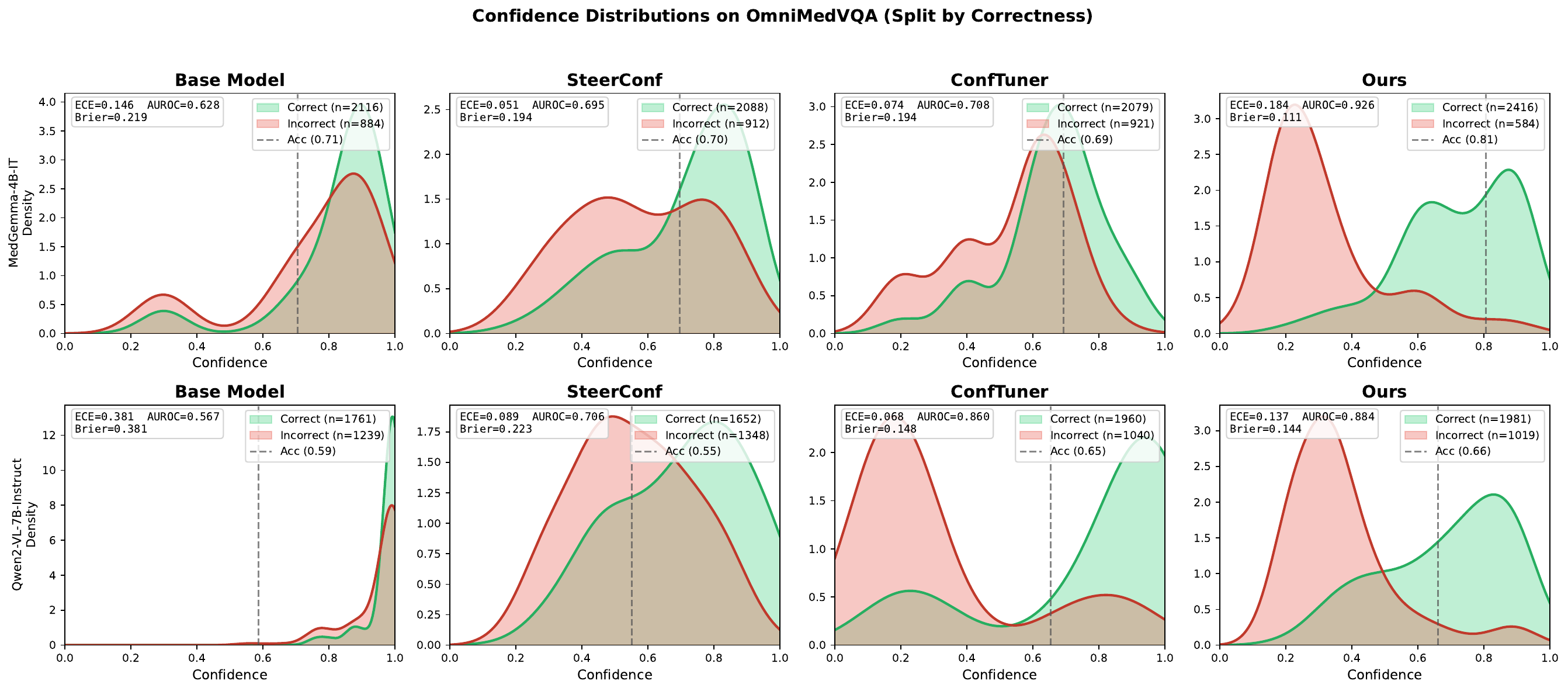}
\caption{Confidence distributions on OmniMedVQA, split by correctness (MedGemma top, Qwen2-VL bottom). Baselines cluster predictions near the dataset accuracy (dashed line), achieving low ECE but overlapping correct and incorrect distributions. Our method separates the two distributions, yielding higher discrimination (AUROC) at the cost of slightly higher ECE.}
\label{fig:kde_compression}
\end{figure*}

\begin{figure*}[t]
\centering
\includegraphics[width=\textwidth]{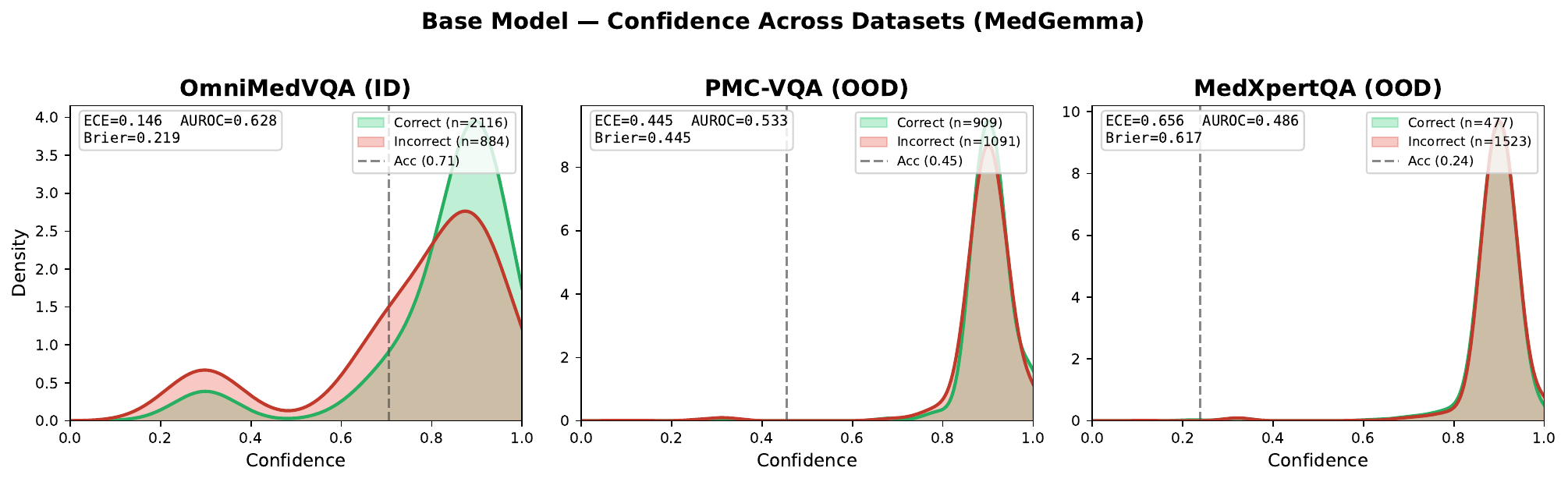}
\caption{Base Model confidence distributions across datasets (MedGemma). The distribution remains concentrated at high confidence regardless of dataset accuracy.}
\label{fig:kde_cross_base}
\end{figure*}

\begin{figure*}[t]
\centering
\includegraphics[width=\textwidth]{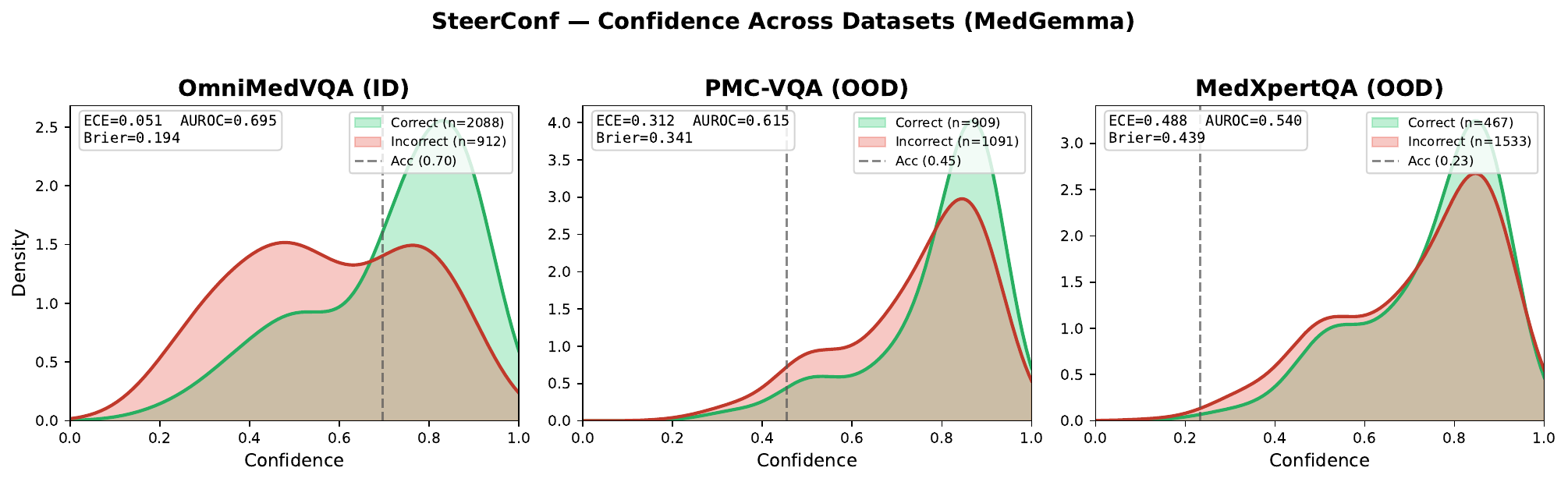}
\caption{SteerConf confidence distributions across datasets (MedGemma). The distribution shape remains largely unchanged despite accuracy dropping from 0.70 to 0.23.}
\label{fig:kde_cross_steerconf}
\end{figure*}

\begin{figure*}[t]
\centering
\includegraphics[width=\textwidth]{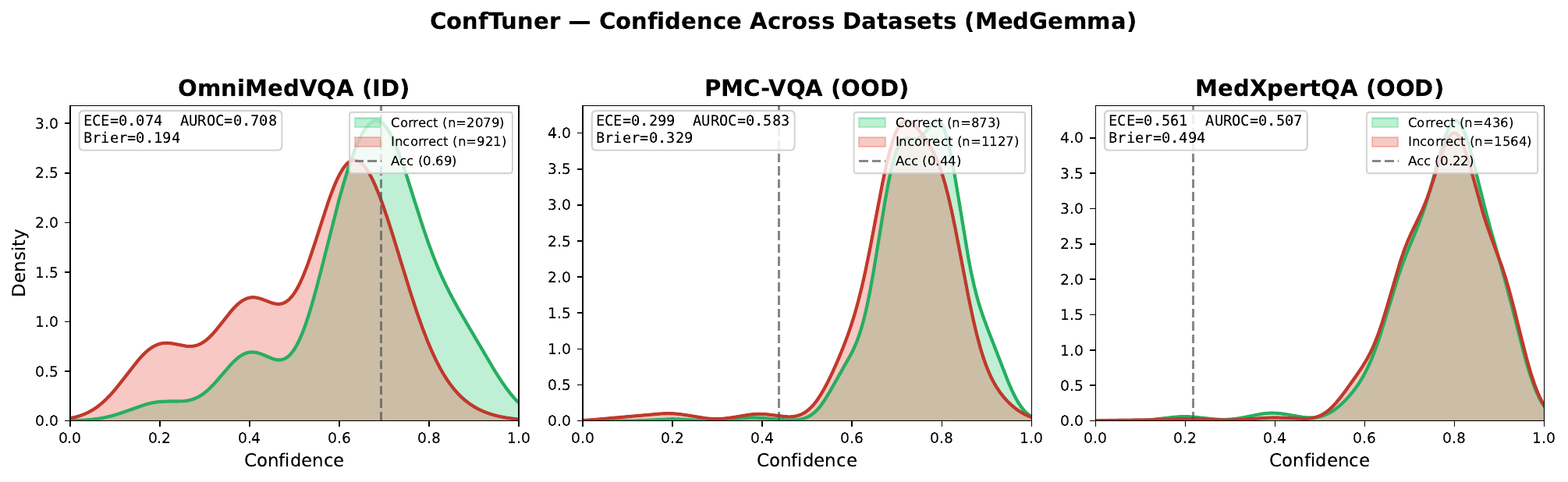}
\caption{ConfTuner confidence distributions across datasets (MedGemma). Confidence remains concentrated at high values even on MedXpertQA where accuracy is 0.22.}
\label{fig:kde_cross_conftuner}
\end{figure*}

\begin{figure*}[t]
\centering
\includegraphics[width=\textwidth]{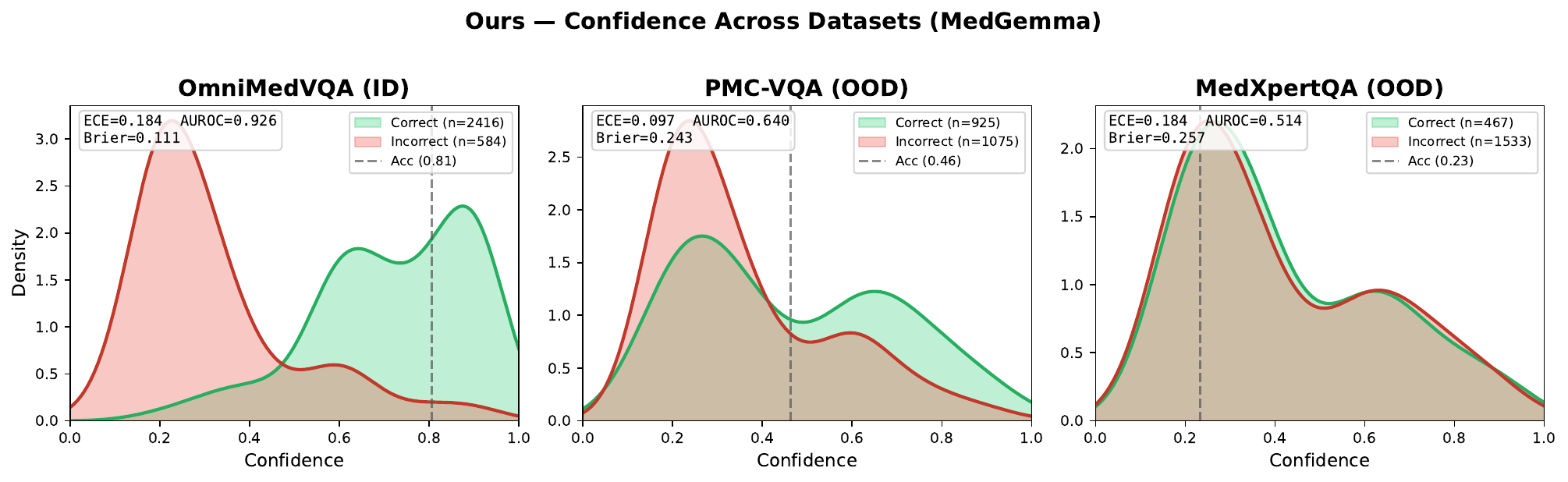}
\caption{Our method's confidence distributions across datasets (MedGemma). The distribution shifts to track dataset accuracy: high confidence on OmniMedVQA (Acc=0.81), moderate on PMC-VQA (Acc=0.46), and low on MedXpertQA (Acc=0.23). Correct and incorrect predictions are well-separated on OmniMedVQA and PMC-VQA; on MedXpertQA, where accuracy approaches random choice, the distributions converge, reflecting the limit of discrimination at this difficulty level.}
\label{fig:kde_cross_ours}
\end{figure*}

Figures~\ref{fig:kde_compression}--\ref{fig:kde_cross_ours} visualize the confidence distributions underlying the metrics reported in Section~\ref{sec:results}.

\paragraph{Per-sample discrimination.} Figure~\ref{fig:kde_compression} compares all methods on OmniMedVQA, splitting predictions by correctness. The base model, SteerConf, and ConfTuner concentrate both correct and incorrect predictions in a similar high-confidence region, producing overlapping distributions. Because OmniMedVQA has high accuracy, this high confidence is close to the dataset mean, which yields low ECE. Our method produces more separated distributions, with correct predictions at higher confidence and incorrect predictions at lower confidence, which yields higher AUROC (Tables~\ref{tab:main_results},~\ref{tab:qwen2_results}).

Table~\ref{tab:ovl} quantifies this separation. We report the mean confidence for correct ($\mu_{\text{corr}}$) and incorrect ($\mu_{\text{incorr}}$) predictions, their gap ($\Delta\mu$), the standard deviation of all scores ($\sigma$), and the Overlapping Coefficient (OVL), which measures the intersection area of the correct and incorrect confidence density functions (lower indicates better separation).

On MedGemma, the base model, SteerConf, and ConfTuner all exhibit high OVL ($\geq$0.698), indicating that their correct and incorrect confidence distributions are largely entangled. The corresponding $\Delta\mu$ values are small (0.077--0.136), confirming that these methods assign similar confidence regardless of correctness. This pattern is consistent with the overconfidence prior of MLLMs: models default to high certainty, and on a high-accuracy benchmark this prior happens to align with the dataset mean, producing low ECE without per-sample informativeness. Our method increases $\Delta\mu$ to 0.406 and reduces OVL to 0.261, achieving clear separation between correct and incorrect densities.

On Qwen2, a similar pattern holds for the base model (OVL\,=\,0.431, $\Delta\mu$\,=\,0.025) and SteerConf (OVL\,=\,0.692). ConfTuner achieves a larger mean gap ($\Delta\mu$\,=\,0.460) than our method ($\Delta\mu$\,=\,0.318), though our method achieves lower OVL (0.356 vs.\ 0.396), indicating tighter separation around the means. Both methods improve over the base model and SteerConf on this in-distribution benchmark. The difference between ConfTuner and our method becomes apparent under distribution shift: ConfTuner's ECE rises from 0.068 on OmniMedVQA to 0.229 on PMC-VQA and 0.360 on MedXpertQA, while our method maintains a narrower range (0.058--0.314), suggesting that the separation learned by our method is more robust to changes in the underlying accuracy regime.

\begin{table}[t]
\centering
\resizebox{\columnwidth}{!}{%
\begin{tabular}{llccccc}
\toprule
\textbf{Arch.} & \textbf{Method} & $\mu_{\text{corr}}$ & $\mu_{\text{incorr}}$ & $\Delta\mu$$\uparrow$ & $\sigma$ & \textbf{OVL}$\downarrow$ \\
\midrule
& Base Model & 0.831 & 0.755 & 0.077 & 0.180 & 0.767 \\
& SteerConf & 0.712 & 0.576 & 0.136 & 0.198 & 0.700 \\
\textbf{MedGemma} & ConfTuner & 0.658 & 0.532 & 0.126 & 0.172 & 0.698 \\
& Ours & 0.723 & 0.317 & 0.406 & 0.236 & 0.261 \\
\midrule
& Base Model & 0.979 & 0.954 & 0.025 & 0.073 & 0.431 \\
& SteerConf & 0.701 & 0.555 & 0.145 & 0.202 & 0.692 \\
\textbf{Qwen2} & ConfTuner & 0.778 & 0.317 & 0.460 & 0.359 & 0.396 \\
& Ours & 0.685 & 0.367 & 0.318 & 0.234 & 0.356 \\
\bottomrule
\end{tabular}%
}
\caption{Confidence distribution statistics on OmniMedVQA, split by correctness. $\Delta\mu$: mean confidence gap between correct and incorrect predictions. $\sigma$: standard deviation of all scores. OVL: Overlapping Coefficient between correct and incorrect confidence densities (lower = better separated).}
\label{tab:ovl}
\end{table}

\paragraph{Adaptivity across difficulty levels.} Figures~\ref{fig:kde_cross_base}--\ref{fig:kde_cross_ours} show how each method's confidence distribution changes across three benchmarks with decreasing accuracy (OmniMedVQA $\rightarrow$ PMC-VQA $\rightarrow$ MedXpertQA). The base model, SteerConf, and ConfTuner produce distributions that are largely fixed: confidence remains concentrated at similar values whether dataset accuracy is 0.70 or 0.23. In a deployment setting where difficulty varies across cases, this rigidity means calibration degrades as soon as the input distribution shifts away from the training regime. Our method's distribution tracks the underlying accuracy, assigning progressively lower confidence as difficulty increases. This is consistent with the narrower ECE spread reported in Section~\ref{sec:results}.

\paragraph{Metric interpretation.} ECE measures whether average confidence matches average accuracy within bins. A method that assigns near-identical confidence to every prediction can achieve low ECE, but such scores do not distinguish correct from incorrect predictions at the sample level. AUROC measures exactly this ranking ability, and the Brier Score captures both calibration and discrimination in a single quantity. In clinical workflows where uncertain predictions should be flagged for review, per-sample ranking is directly actionable, which motivates optimizing for discrimination alongside calibration.

\subsection{Visual Evidence Removal Design}
\label{sec:a_black_image}

The $2\times2$ factorial design (Section~\ref{sec:perturbation}) requires a method for removing visual evidence from the input. Noise-based alternatives such as Gaussian blur or mean pixel padding do not guarantee removal of diagnostic content: determining whether a given perturbation intensity actually destroys the clinically relevant signal requires domain expertise or expert annotation, and the answer varies across the diverse imaging modalities in our training data (fundus photography, dermatoscopy, X-ray, ultrasound, histopathology). A black image guarantees complete evidence removal with no domain expertise and no per-modality tuning, making it a simple, modality-agnostic solution. The fractional accuracy under each condition is then measured empirically, so the training signal reflects how the model actually behaves under evidence removal rather than assumptions about the perturbation.

Tables~\ref{tab:black_image_medgemma} and~\ref{tab:black_image_qwen} show representative outputs under the black-image condition. Both models hallucinate visual findings and produce coherent medical reasoning based on language priors. The outputs are structured and fluent rather than degenerate, indicating that the model falls back on textual priors rather than producing artifacts from out-of-distribution visual encoder activations. In some cases the model also answers correctly under the black-image condition, which is expected: certain questions are solvable from language priors alone (e.g., common medical knowledge or option elimination), and the fractional accuracy $\hat{a}_{\text{black}}$ captures this. The black image serves as a proxy for probing what evidence the model relies on when forming its answer. Alternative proxies such as learned occlusion masks or domain-aware degradation may provide a more targeted signal and are a direction for future work.

\begin{table}[t]
\centering
\small
\begin{tabular}{p{0.06\columnwidth} p{0.88\columnwidth}}
\hline
\# & \textbf{Question / Model Output} \\
\hline
1 & \textit{``What is the specific abnormality present in this image?''} (A=Optic neuritis, B=Conjunctivitis, C=Proliferative diabetic retinopathy, D=Macular degeneration) \newline \textbf{Correct:} C \quad \textbf{Model:} ``The image shows a fundus photograph of the retina. There are no obvious signs of optic neuritis, conjunctivitis, macular degeneration, or proliferative diabetic retinopathy.'' $\rightarrow$ \textbf{A} \\
\hline
2 & \textit{``What is the specific condition depicted in this image?''} (A=Rubella, B=Measles, C=Cowpox, D=Chickenpox) \newline \textbf{Correct:} C \quad \textbf{Model:} ``The image shows a clear, well-defined, and relatively uniform area of skin discoloration, which is a hallmark of a skin condition.'' $\rightarrow$ \textbf{A} \\
\hline
3 & \textit{``What is the diagnosis of the cancer shown in the image?''} (A=Renal cell carcinoma of the left hilum T2 N1 M0 Stage II, B=Adenocarcinoma, C=Squamous cell carcinoma, D=Small cell carcinoma) \newline \textbf{Correct:} C \quad \textbf{Model:} ``The image shows a solid, homogenous mass in the hilum. The most likely diagnosis is renal cell carcinoma based on the location and appearance.'' $\rightarrow$ \textbf{A} \\
\hline
\end{tabular}
\caption{MedGemma-4B-IT outputs under the black-image condition. The model hallucinates visual findings and reasons from language priors, producing coherent but incorrect answers.}
\label{tab:black_image_medgemma}
\end{table}

\begin{table}[t]
\centering
\small
\begin{tabular}{p{0.06\columnwidth} p{0.88\columnwidth}}
\hline
\# & \textbf{Question / Model Output} \\
\hline
1 & \textit{``What can be observed in this image?''} (A=Joint dislocation, B=Enlarged lymph node, C=Soft tissue fluid, D=Nerve inflammation) \newline \textbf{Correct:} C \quad \textbf{Model:} ``The image shows a circular, well-defined area with increased signal intensity, which is characteristic of an enlarged lymph node.'' $\rightarrow$ \textbf{B} \\
\hline
2 & \textit{``What is the noticeable condition depicted in this image?''} (A=Carpal tunnel syndrome, B=Achilles pathology, C=Arthritis, D=Tendonitis) \newline \textbf{Correct:} B \quad \textbf{Model:} ``The image shows a hand with a prominent swelling and tenderness, which is characteristic of a tendonitis.'' $\rightarrow$ \textbf{D} \\
\hline
3 & \textit{``What is the organ depicted in this ultrasound image?''} (A=Intestine, B=Testis, C=Kidney, D=Pancreas) \newline \textbf{Correct:} C \quad \textbf{Model:} ``The image shows a structure with a central hypoechoic area surrounded by a more echogenic rim, which is characteristic of the testis.'' $\rightarrow$ \textbf{B} \\
\hline
\end{tabular}
\caption{Qwen2-VL-7B outputs under the black-image condition. Same pattern: the model generates plausible medical descriptions but answers incorrectly, relying on language priors.}
\label{tab:black_image_qwen}
\end{table}

\subsection{Calibrated Confidence in Practice}
\label{sec:a_deployment}

Standard calibration metrics such as ECE, Brier score, and AUROC quantify statistical properties of confidence distributions.
To examine what these properties imply in a more applied context, we consider a simplified deployment scenario: a clinical setting where an MLLM assists with diagnostic cases.
A deployed model encounters both familiar and novel cases.
OmniMedVQA represents in-distribution cases, analogous to a hospital's own historical records on which the model was trained.
PMC-VQA and MedXpertQA represent out-of-distribution cases, analogous to new or previously unseen case types.
In addition to the distribution shift, these datasets vary in relative difficulty for the model: OmniMedVQA is the easiest (highest accuracy), PMC-VQA is moderate, and MedXpertQA is the hardest, collectively approximating the range of case complexity a deployed system may encounter.

Before deployment, the system must commit to a confidence threshold that governs its behavior.
The choice of threshold determines which cases the model acts on and which it defers.
We examine two paradigms below, both dependent on this threshold and therefore on the quality of the underlying confidence estimates.
This analysis examines whether the calibration improvements reported in Section~\ref{sec:results} translate to observable differences in simplified deployment scenarios, bridging the gap between aggregate metrics and deployment-relevant behavior.
As discussed in Section~\ref{sec:a_experimental_details}, the multiple-choice format enables reliable automated correctness assessment, which is necessary for the analysis below.

\subsubsection{AI-Assisted Clinician}
\label{sec:a_ai_assisted}

In this paradigm the clinician remains the primary decision-maker and the model serves as an advisory tool.
For such a system to be practical, the model should provide recommendations selectively, only when it is sufficiently confident, rather than on every case.
Unrestricted recommendations increase clinician workload: each suggestion must be verified, and incorrect confident suggestions may introduce bias or require additional effort to rule out.
An ideal advisory model would therefore achieve high precision in its recommendations, offering fewer but more reliable second opinions.
Calibrated confidence could enable this behavior by allowing the system to suppress recommendations below a confidence threshold, surfacing only cases where the model's prediction is likely correct.
Such selective recommendations could help clinicians catch diagnostic errors, obtain a reliable second opinion on ambiguous cases, or prioritize cases that warrant closer examination.

\begin{figure*}[t]
\centering
\includegraphics[width=\textwidth]{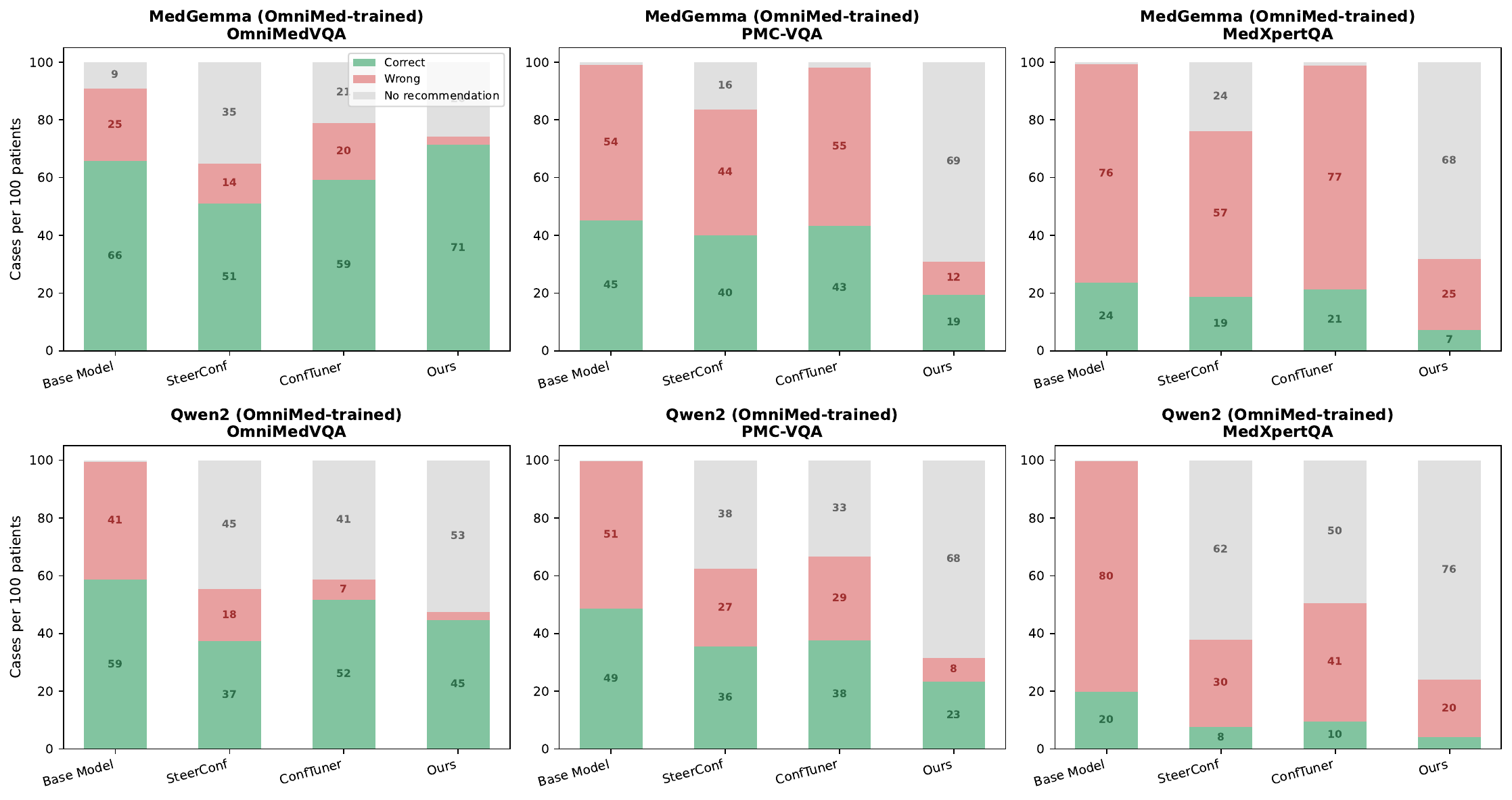}
\caption{Recommendation outcomes per 100 patients at confidence threshold 6. Green: correct recommendations. Pink: wrong recommendations. Gray: model abstains. Our method (rightmost bar in each group) produces fewer wrong recommendations by abstaining when uncertain, while baselines recommend indiscriminately regardless of dataset difficulty.}
\label{fig:adaptive_coverage}
\end{figure*}

Figure~\ref{fig:adaptive_coverage} quantifies this effect at confidence threshold 6 (on a 1--10 scale), chosen as an illustrative high-precision operating point: an advisory system should only surface recommendations it is reasonably certain about.
Each bar represents 100 patients and is divided into correct recommendations (green), wrong recommendations (pink), and cases where the model abstains (gray).
Figure~\ref{fig:error_catch} later confirms that this pattern holds across the full threshold range.

Baselines assign high confidence scores regardless of dataset difficulty.
On MedXpertQA, where overall accuracy is approximately 20\%, the base model still recommends on nearly all cases, producing 76 wrong recommendations per 100 patients for MedGemma and 80 for Qwen2.
Our method reduces these to 25 and 20 respectively by abstaining on cases where its confidence is low.
The abstention is not uniform: on OmniMedVQA, where accuracy is higher, our method recommends on approximately 74\% of cases (MedGemma) and 47\% (Qwen2) with 71 and 45 correct recommendations, while producing only 3 wrong recommendations in both cases.

This behavior emerges directly from calibration.
Because our training procedure encourages confidence scores that track actual correctness likelihood, the model naturally scales back its recommendations on harder datasets without any explicit abstention mechanism.
An overconfident model lacks this adaptivity: it recommends with similar frequency on OmniMedVQA and MedXpertQA despite a threefold difference in accuracy, exposing clinicians to a high volume of incorrect suggestions on difficult cases.
In practice, each wrong recommendation requires the clinician to identify the error and override the suggestion, adding verification effort that may exceed the time saved by correct recommendations.
A calibrated model that abstains on uncertain cases reduces this overhead, providing fewer but more time-efficient second opinions.

\subsubsection{Clinician-Assisted AI}
\label{sec:a_clinician_assisted}

In this paradigm the model operates semi-autonomously, handling the majority of cases and routing predictions with confidence below a threshold to a human reviewer.
This setup is relevant to high-throughput settings where case volume exceeds clinician capacity, and the model serves as a first-pass filter.
For such a safety gate to be functional, low confidence must correspond to cases that actually need review.
An overconfident model assigns high scores indiscriminately, rendering the threshold ineffective, and errors pass through undetected regardless of the chosen operating point.

\begin{figure*}[t]
\centering
\includegraphics[width=\textwidth]{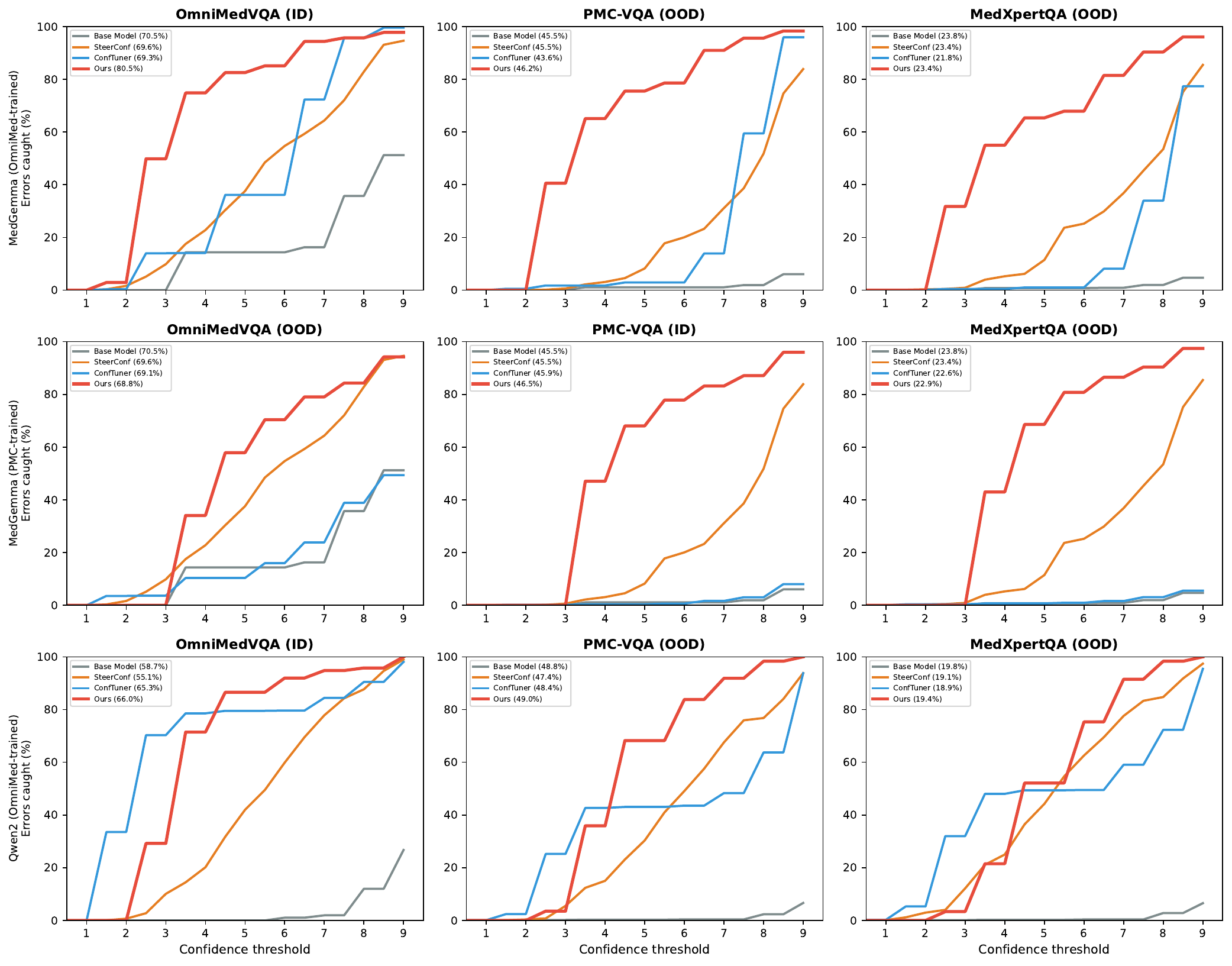}
\caption{Error catch rate vs.\ confidence threshold. Each curve shows the fraction of model errors with confidence below the threshold. Percentages in the legend denote each method's accuracy on that dataset. Rows correspond to training configurations; columns to evaluation datasets (ID or OOD).}
\label{fig:error_catch}
\end{figure*}

Figure~\ref{fig:error_catch} plots the fraction of model errors that fall below each confidence threshold across the full range (1--9), showing that the pattern from Section~\ref{sec:a_ai_assisted} is not specific to a single threshold.
On MedGemma (rows 1--2), our method reaches higher error catch rates across the full threshold range in both ID and OOD settings.
On Qwen2 (row 3), ConfTuner is competitive on the in-distribution dataset at lower thresholds, though our method catches a larger fraction at higher thresholds.
The difference is most visible on out-of-distribution datasets, where baseline curves remain near zero at moderate thresholds because overconfident scores prevent the safety gate from triggering.

To illustrate with a concrete operating point, we examine threshold 3, a high-recall setting where the safety net casts a wider net to catch as many errors as possible.
On MedGemma (OmniMed-trained) at this threshold, our method flags 10.6\% of cases with 91.5\% precision (i.e., 9 out of 10 flagged cases are genuine errors) and catches 49.8\% of all errors.
At the same threshold, SteerConf flags 4.6\% at 65.7\% precision catching 9.9\%, and the base model flags 0.1\% catching effectively nothing.
The pattern is consistent across architectures: on Qwen2 at threshold 3, our method achieves 92.8\% precision while catching 29.2\% of errors.
Overall, the improved calibration from our training procedure translates to a more practical safety gate, catching a larger share of errors at lower review cost compared to baselines, and suggests higher deployment readiness for threshold-based human-in-the-loop systems.

While this analysis operates within the multiple-choice evaluation setting, it demonstrates a behavioral prerequisite for clinical deployment: a model must be able to reliably modulate its output, abstaining on cases it is likely to answer incorrectly and engaging on cases where it can contribute.
Our calibration framework provides the confidence signal necessary to enable this behavior.

\end{document}